\documentclass[manuscript]{acmart}%,authorversion

\usepackage[utf8]{inputenc} % allow utf-8 input
\usepackage[T1]{fontenc}    % use 8-bit T1 fonts
\usepackage{hyperref}       % hyperlinks
\usepackage{url}            % simple URL typesetting
\usepackage{booktabs}       % professional-quality tables
\usepackage{amsfonts}       % blackboard math symbols
\usepackage{nicefrac}       % compact symbols for 1/2, etc.
\usepackage{microtype}      % microtypography
\usepackage{xcolor}         % colors
\usepackage{graphicx}
\usepackage{bbm}
\usepackage{amsmath}
\usepackage{subcaption}

\usepackage{placeins}

\newlength\stextwidth
\newcommand\makesamewidth[3][c]{%
  \settowidth{\stextwidth}{#2}%
  \makebox[\stextwidth][#1]{#3}%
}

\usepackage[inline]{enumitem}

\usepackage{mathtools}
\newtheorem{theorem}{Theorem}[section]

\newtheorem{definition}[theorem]{Definition}

\AtBeginDocument{%
  }

\acmYear{2024}\copyrightyear{2024}
\acmConference[ACM FAccT '24]{ACM Conference on Fairness, Accountability, and Transparency}{June 3--6, 2024}{Rio de Janeiro, Brazil}
\acmBooktitle{ACM Conference on Fairness, Accountability, and Transparency (ACM FAccT '24), June 3--6, 2024, Rio de Janeiro, Brazil}
\acmDOI{10.1145/3630106.3658989}
\acmISBN{979-8-4007-0450-5/24/06}

\begin{document}

\title[Equalised Odds Is Not Equal Individual Odds]{Equalised Odds Is Not Equal Individual Odds:\\ Post-processing for Group and Individual Fairness}


\author{Edward Small}
\affiliation{%
  %\institution{Royal Melbourne Institute of Technology}
  \institution{RMIT University}
  \city{Melbourne}
  \country{Australia}}
\email{edward.small@student.rmit.edu.au}
\orcid{0000-0002-3368-1397}

\author{Kacper Sokol}
\affiliation{%
  \institution{ETH Zurich}
  \city{Zurich}
  \country{Switzerland}}
\email{kacper.sokol@inf.ethz.ch}
\orcid{0000-0002-9869-5896}

\author{Daniel Manning}
\affiliation{%
  %\institution{Royal Melbourne Institute of Technology}
  \institution{RMIT University}
  \city{Melbourne}
  \country{Australia}}
\email{daniel.manning@student.rmit.edu.au}
\orcid{0009-0007-1144-3121}

\author{Flora D.\ Salim}
\affiliation{%
  %\institution{University of New South Wales}
  \institution{UNSW Sydney}
  \city{Sydney}
  \country{Australia}}
\email{flora.salim@unsw.edu.au}
\orcid{0000-0002-1237-1664}

\author{Jeffrey Chan}
\affiliation{%
  %\institution{Royal Melbourne Institute of Technology}
  \institution{RMIT University}
  \city{Melbourne}
  \country{Australia}}
\email{jeffrey.chan@rmit.edu.au}
\orcid{0000-0002-7865-072X}

\renewcommand{\shortauthors}{Small et al.}

\begin{abstract}
Group fairness is achieved by equalising prediction distributions between protected sub-populations; individual fairness requires treating similar individuals alike. These two objectives, however, are incompatible when a scoring model is calibrated through \emph{discontinuous} probability functions, where individuals can be randomly assigned an outcome determined by a fixed probability. This procedure may provide two similar individuals from the same protected group with classification odds that are disparately different -- a clear violation of individual fairness.
Assigning unique odds to each protected sub-population may also prevent members of one sub-population from ever receiving the chances of a positive outcome available to individuals from another sub-population, which we argue is another type of unfairness called \emph{individual odds}. We reconcile all this by constructing continuous probability functions between group thresholds that are constrained by their Lipschitz constant. Our solution preserves the model's predictive power, individual fairness and robustness while ensuring group fairness. 
\end{abstract}





\maketitle

\section{Introduction}
\label{Introduction}

Predictive models that output a score or probability for a multi-dimensional input, i.e., scoring functions, are a common tool in automated decision-making~\citep{BayesOptimal, binaryclass}. Binary classification is a popular realisation of this paradigm, where a threshold is placed on a score to produce a decision; among others, it can be found in school examinations where individual answers are condensed into a grade that translates to a pass/fail mark~\citep{passfail}, or banking where the history of personal finances is compressed into a credit score that captures one's likelihood of defaulting on a loan~\citep{MARKOV2022180}. Many such applications, especially in high stakes domains like healthcare, finance and judiciary, are coming under increased scrutiny given their potential harm to society -- predictive models deployed in these contexts are expected to be accurate, robust, fair and explainable. These four desiderata, however, are often at odds. Improving utility, i.e., predictive power, of a model may entail increasing its complexity at the expense of interpretability and robustness, e.g., due to overfitting~\citep{pmlr-v162-yu22b, NEURIPS2020_b1adda14}. Similarly, equalising errors between protected groups to ensure fairness may require sacrificing utility and impairing other notions of fairness~\citep{dutta20a, Wang2021UnderstandingAI}.% counteracting

In this paper we focus on the latter scenario, where (protected) sub-populations are treated differently, thus unfairly, due to persistent historical biases~\citep{NEURIPS2018_1f1baa5b}, training data under-representation~\citep{underbias} and greedy optimisation of an objective function. Correcting for these biases is often challenging as it requires detailed knowledge of the data domain and the input space. One popular solution to this problem, which we study here, is threshold optimisation under fairness constraints when dealing with multiple protected groups. This method relies on calculating unique decision functions based on scores for each protected group in order to satisfy a given fairness constraint, e.g., demographic parity~\citep{vogel2021learning}.

\begin{figure}[t]
%\begin{center}
\centering
\includegraphics[width=0.85\columnwidth]{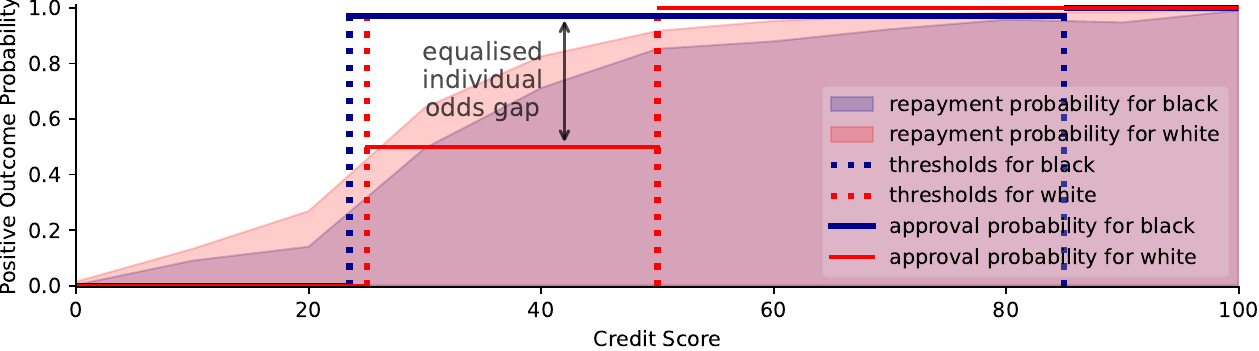}
\caption{Two-threshold fixed randomisation~\citep{NIPS2016_9d268236} applied to probabilities (y-axis) output by a loan repayment classifier built upon credit scores (x-axis). It satisfies equalised odds for the binary protected attribute \textit{race} (black and white) by using it to assign approval probabilities, but results in discontinuities that violate individual fairness and create a gap between group-specific \emph{individual odds}.}%  unique to each group
\label{multi-threshold}
%\end{center}
\end{figure}

We re-examine this approach, as finding a set of thresholds for a given score function that satisfy multiple fairness constraints -- such as equalised odds~\citep{eo} -- is often impossible if only using a collection of single thresholds. Instead, a decision function that is optimal with respect to a definition of group fairness selected by the model owner is derived directly from the scoring function using a pair of thresholds for each group. Outputs that fall between the thresholds are allocated a random decision based on a fixed probability parameter -- a procedure called \textit{fixed randomisation} -- which, while effective, exhibits a number of shortcomings demonstrated by Figure~\ref{multi-threshold} and discussed later in Section~\ref{postproc}. Using a fixed randomisation parameter is suboptimal for both the entities that create the model (owners) and those whose case is being decided by the model (users) because:
\begin{enumerate}%[label=(\arabic*)]
\item if the scoring function is accurate, the decision function cannot leverage this in the intervals between the thresholds;
\item even if the scoring function is individually fair, the step-based decision function is not, e.g., users whose scores are just under a threshold are treated very differently to those who are barely above it, despite their scores being similar (commonly referred to as the \textit{threshold effect}~\cite{threshold_effect}); and
\item users from one protected group may be unable to access the odds of positive classification offered to another group (equalised \emph{individual odds} unfairness).
\end{enumerate}

Consider the fixed randomisation solution shown in Figure~\ref{multi-threshold}, which satisfies equalised odds for the two values of the protected attribute \emph{race} in a loan allocation setting using fixed randomisation. For example, a \emph{white} user with a credit score of $49.5$ is assigned the same odds (50\%) of receiving a loan as a \emph{white} user whose credit score is $25$ despite the latter being $6.6$ times more likely to default than the former. This stands in stark contrast to a \emph{white} user with a credit score of $24.5$ (just below the threshold) who has no chance of getting a loan despite being only $1.03$ times more likely to default than the aforementioned \emph{white} user with a credit score of $25$.
This case study illustrates that an increase in credit score -- and therefore an increase in the likelihood of repaying a loan -- is not reflected in the final decision for all scores except at the thresholds. In addition, while some \emph{white} users have a 50\% chance of receiving a loan and some \emph{black} users have a 97.2\% chance, these success odds are never offered to the other group; therefore, a \emph{white} user will never be given a 97.2\% chance of receiving a loan and vice versa. This disparity motivates a new notion of fairness -- called \emph{equalised individual odds} -- which we outline in Definition~\ref{individualodds} in Section~\ref{postproc}.

We address these shortcomings by deriving a set of closed-form, continuous, monotonic functions (shown later in Figure~\ref{curves}) parameterised only by the thresholds and a probability parameter, making them easy to compute (Section~\ref{systems}). We show that these functions are constrained via a maximum derivative, preventing a change in score leading to a large shift in classification odds and thus maintaining individual fairness and softening the threshold effect (Section~\ref{sec::lip}). Our approach enables the model owners to prioritise users with higher scores, better honouring the underlying score distribution as well as improving the transparency of the process. These properties incentivises users to increase their score as such an action improves their odds of a positive outcome -- see Figure~\ref{multi-threshold_cont} for a direct comparison to Figure~\ref{multi-threshold}. We analyse our method in two case studies -- through the lens of credit scoring for loan allocation in Section~\ref{sec:creditriskstudy}, and risk of recidivism in Section~\ref{sec:compasstudy}. For the credit scoring case study we seek equalised odds across the four values of the \emph{race} attribute -- non-Hispanic white (white), black, Hispanic and Asian -- found in the \emph{2003 TransUnion TransRisks Scores} (CreditRisk) data set~\citep{feds}, whereas for the recidivism case study we enforce equalised odds across a combination of two \emph{races} -- Caucasian and African America -- and two \emph{sexes} -- male and female --  found in the \emph{2016 ProPublica Recidivism Risk Score} (COMPAS) data set~\citep{feds}. In both cases, we show that individual fairness is improved while group fairness and accuracy are preserved. %
 % captures
In summary, our contribution is threefold:
\begin{enumerate*}[label=(\arabic*)]
    \item we demonstrate that fixed randomisation for group fairness violates individual fairness;
    \item we  derive a set of closed-form, continuous and monotonic probability functions; and
    \item we show that these continuous curves preserve group fairness and improve performance while adhering to the constraint imposed by individual fairness.
\end{enumerate*}

\section{Preliminaries}
\label{sec:prelim}

\subsection{Notation}
We assume that the scalar scoring function $g:\mathcal{X}\mapsto \mathcal{R}$ takes individual instances and outputs a score $\mathcal{R}\subseteq\mathbb{R}$; $h:\mathcal{R} \mapsto \mathcal{Y}$, where $\mathcal{Y}\equiv\{0,1\}$, is an arbitrary, possibly stochastic, binary decision function on $\mathcal{R}$ that maps the scores $R$ to predicted classes $\widehat{Y}$ according to a predetermined probability distribution $\mathbb{P}\{\widehat{Y}=1\big\vert R=r\}$. Lower case letters denote an individual instance from a sample, e.g., $\mathbf{x}$ is an instance in $X$. Functions denoted by Greek letters, such as $\zeta:\mathcal{R}\mapsto \mathcal{I}$ where $\mathcal{I}\equiv[0,1]$, parameterise this probability based on scores, e.g., according to the Bernoulli distribution $h(r)\sim B\big(1, \zeta(r)\big)$. Effectively, $h(r)$ is the probability that $\widehat{Y}=1$ for $R=r$. Alternatively, for deterministic behaviour $\zeta$ can be defined by a single threshold $t\in \mathcal{R}$, where a score $r \geq t$ yields $h(r)=1$ and $r < t$ yields $h(r)=0$; this behaviour can be captured by the indicator function
\begin{equation*}
    \zeta(r) = \mathbbm{1}_{t}(r) = 
        \begin{cases}
            0 &\textrm{if} \;\; r < t \\
            1 &\textrm{if} \;\; r \geq t
        \end{cases} 
            \text{~.}
\end{equation*}
One common realisation of this thresholding function is a binary probabilistic classifier, where $\mathcal{R}\equiv\mathcal{I}$ and $t=0.5$. We therefore define the final \textit{decision function} $f_h:\mathcal{X}\mapsto\mathcal{Y}$ as the composition $f_h=h \circ g$, where the subscript on $f$ indicates the composition function $h$ on the scoring function $g$.
Additionally, capital letters refer to samples from spaces, such that $X$ is a sample from the space $\mathcal{X}$; $g(X)=R$ are the corresponding scores calculated by $g$ for the sample $X$; $f(X)=\widehat{Y}$ are classes predicted for all instances in the sample $X$; and $Y$ captures their ground truth labels. We denote the protected attribute as $A$, and consider the joint distribution $(R, A, Y)$. We make no assumptions on the type or shape of $\mathcal{X}$, nor on the construction of $g$ (the behaviour of which is discussed in Section~\ref{postproc}).

\subsection{Distance and Similarity Measures}
\label{SimilarIndividuals}

Defining ``similar individuals'' can be challenging and is deeply rooted in the landscape and shape of the input space, the complexity of the problem, and the density and distribution of the training data $X$ within the space. Distances on metric spaces, regardless of their definition, must follow a set of axioms (outlined in Appendix~\ref{ap:dist}). This problem is also not strictly mathematical and depends highly on the context. Additionally, discrete or categorical data can be difficult to quantify and compare; for example, in a feature space of size $N$, how different is an unmarried individual from a married person, all other things being equal? One could argue that its importance depends on the size of $N$ -- a large value of $N$ can dilute the importance of each individual feature. If we are trying to predict whether an individual has any children, however, this feature is of high importance regardless of the size of $N$. To best capture such dependencies, we can employ similarity graphs or bespoke distance metrics chosen based on the problem definition and the data set at hand.

Using tailor-made definitions of similarity, nonetheless, poses two issues:
\begin{enumerate*}
\setlength\itemsep{0em}
    \item it makes it difficult to compare results between experiments; and
    \item the results are subject to the quality of the metric and its suitability for the problem at hand.
\end{enumerate*}
We operate under the assumption that model inputs are inaccessible (simulating scenarios where data and model parameters are protected and/or private), thus we are only given scores, values of the protected attribute and the label (ground truth).
For our work we therefore rely on generic distance metrics such as Euclidean, Hamming and Gower's distances.
Note that we assume that changing the protected class $A$ for an individual is too large of a change to label the two instances as similar since this alteration entails using a different set of thresholds and probabilities in the final decision function. Examples of classical distance functions are presented in Appendix~\ref{ap:dist}.

\subsection{Related Work}
\label{sec:fair}

%\subsection{Group and Individual Fairness}
%\label{sec:fair}
%\paragraph{Group Fairness}%
Group and individual fairness are two commonly considered categories~\citep{ivg}. 
\textbf{Group fairness} focuses on the statistical difference in outcomes between sub-populations determined by the values of a protected attribute $A$~\citep{statsoutcome}. The type of statistical outcome that a model owner may want to focus on is domain-specific, but measures closer to $0$ are more desirable as this indicates no statistical difference between two groups. For a simple case of a binary protected feature $A=\{a,a^\prime\}$ where $a \cap a^\prime=\emptyset$, we can further differentiate two types of group fairness~\citep{clarify}:
\begin{description}
    \item [Outcome] Predictions are equalised in a set way across groups, e.g., demographic parity~\citep{agarwal2018a}:
    \begin{equation*}
        \left\lvert \mathbb{P}\{\widehat{Y}=1\big\vert A=a\} - \mathbb{P}\{\widehat{Y}=1\big\vert A=a^\prime\} \right\rvert=0 \text{~.}
        %\label{demopar}
    \end{equation*}
    An example of demographic parity may be in school admissions~\citep{school}, where the distribution of admitted students should represent the distribution of the applicants for each value of $A$ (i.e., if applicants are $50\%$ male and $50\%$ female, admissions should reflect this pattern).
    
   \item [Error Distribution] (In)correct classifications should be equalised in a predetermined way, e.g., using false negative rate:
    \begin{equation*}
        \left\lvert  \mathbb{P}\{\widehat{Y}=0\big\vert A=a,Y=1\} - \mathbb{P}\{\widehat{Y}=0\big\vert A=a^\prime,Y=1\} \right\rvert=0
        \text{~.}
        \label{FP}
    \end{equation*}
    An example of equalising false negative rate may be in the medical field, where false negatives could have dire consequences for a patient. Erring on the side of caution equally for all groups is therefore more preferable, up to a certain cost~\citep{health}.
\end{description}
There are many ways in which group fairness can be operationalised, with different tasks and domains requiring a specific constraint or a mixture thereof.
In this paper, we mainly consider one of the strongest fairness constraints called \textit{equalised odds}~\citep{NEURIPS2020_03593ce5}, which is outlined in Definition~\ref{def::EO}.

\begin{definition}[Equalised Odds]
\label{def::EO}
    A decision function $f:\mathcal{X}\mapsto\mathcal{Y}$ satisfies equalised odds with respect to a protected attribute $A$ if false positives and true positives are independent of the protected attribute:
    \begin{equation*}
    \begin{aligned}
        \left\lvert
        \mathbb{P}\{\widehat{Y}=1 \big\vert A = a, Y=y\} - \mathbb{P}\{\widehat{Y}=1 \big\vert A = a^\prime, Y=y\}
        \right\rvert=0
        \;\;
        \forall y\in\mathcal{Y}
        \;\; \forall a,a^\prime\in A
        \;\;
        a\neq a^\prime
        \text{~.}
        \end{aligned}
        %\label{EO}
    \end{equation*}
    %where $a\neq a^\prime$.
\end{definition}

A large portion of fairness research in machine learning therefore focuses on equalising outcomes and errors between users who belong to different protected groups, such as race or sex~\citep{biassurvey}. There are three distinct areas where fairness can be injected into a data modelling pipeline:
\begin{description}
% \emph{pre-processing}
\item [pre-processing] transforms the underlying training data such that signals and cross-correlations causing bias and discrimination are weakened~\citep{pre};
% \emph{in-processing}
\item [in-processing] incorporates fairness constraints directly into the optimisation objective~\citep{in}; and
% \emph{post-processing}
\item [post-processing] alters the output of a decision-making process to mitigate bias of the underlying (fixed) model~\citep{8682620}.
\end{description}
A variety of methods is needed as even when
the scoring function is trained as ``unaware''~\citep{unaware}, and as such has no knowledge of the value of the protected class $A$, $f$ can still become unfair. For example, the ground truth $Y$ may be correlated with $A$ due to historical biases, some features in $X$ may act as a proxy, or the distribution/behaviour of some features in $X$ may be different between sub-populations, causing a predictive model to under-perform for under-represented groups.
A different strand of work looks into fair data collection~\citep{FAIR} and feature selection~\citep{feature} as well as fair learning procedures, e.g., adversarial learning~\citep{adverse}. In this paper, we focus on a popular class of post-processing methods known as \emph{threshold optimisation}. Our work builds directly upon the foundational method introduced by \citet{NIPS2016_9d268236} by expanding and improving it along multiple dimensions.

%\paragraph{Individual Fairness} 
%\label{IF}
A slightly more nuanced view on fairness is the notion of ``treating similar individuals
similarly'', known as \textbf{individual fairness}~\citep{NEURIPS2021_d9fea4ca}. In short, we look to impose a constraint on the distance between any two points (individuals) in the input space against their distance in the output space~\citep{unaware}. We measure distance or similarity using distance functions $d$ on the input ($d_\mathcal{X}$) and output ($d_\mathcal{R}$) spaces: 
\begin{equation}
    d_\mathcal{R} \left( g(\mathbf{x}_1), g(\mathbf{x}_2) \right) \leq L_\mathcal{X} d_\mathcal{X}(\mathbf{x}_1,\mathbf{x}_2) \implies \frac{d_\mathcal{R}(r_1, r_2)}{d_\mathcal{X}(\mathbf{x}_1,\mathbf{x}_2)} \leq L_\mathcal{X} \qquad \forall\mathbf{x}_1,\mathbf{x}_2\in\mathcal{X}
    \text{~,}
    \label{lip}
\end{equation}
where $r_k = g(\mathbf{x}_k)$ and $L_\mathcal{X}\geq0$ is a \textit{Lipschitz constant} (refer to Section~\ref{SimilarIndividuals} for a discussion of distance metrics). The Lipschitz constant describes the maximal difference of the distance between two values in the input space with their corresponding distance in the output space. Limiting $L_\mathcal{X}$ is usually done with a smoothing process, e.g., manifold regularisation~\citep{JMLR:v7:belkin06a}, or by constraining the optimisation of $g$ subject to a condition on the size of $L_\mathcal{X}$. The concept is to assume that individuals with similar features (small $d_\mathcal{X}$) should appear close together in the output space (small $d_\mathcal{R}$). Therefore, limiting the rate at which $g$ can change (i.e., its differential) in densely populated areas of the feature space can force $g$ to be smoother, hence more fair.
$d_\mathcal{X}$ can be as simple as Gower's distance for mixed categorical and numerical features (see Appendix~\ref{ap:dist}), but ideally should be chosen appropriately for the problem at hand;
since $\mathcal{R}$ is scalar, we can define %:
$
%\begin{equation*}
    d_\mathcal{R} \left( g(\mathbf{x}_1), g(\mathbf{x}_2) \right) = \lvert r_1 - r_2 \rvert 
%    \text{~.}
$.
%end{equation*}

\section{Post-processing for Fairness with Fixed Randomisation}
\label{postproc}

In general, we expect that the higher the score $r$ output by a scoring function $g$ used for a predictive task, the ``better'' the outcome. This property is known as \emph{positive orientation}, with \emph{negative orientation} describing the opposite behaviour~\citep{scores2}. For example, if $\mathbf{x}_1$ and $\mathbf{x}_2$ are randomly drawn instances from $X$ used to calculate credit scores, and $\mathbf{x}_1$ has a higher credit score than $\mathbf{x}_2$ -- i.e., $g(\mathbf{x}_1) > g(\mathbf{x}_2)$ -- this relation implies that $\mathbf{x}_1$ is more likely to have healthier spending habits, thus making this person more likely to repay a loan (see Figure~\ref{credit_pmf} in Appendix~\ref{ap:prob_func}). This does not need to be strictly true for all values, but should hold in general. In other words, we expect the receiver operating characteristic (ROC) curve -- which expresses the (\emph{false positive}, \emph{true positive}) rates at different thresholds on $R$ -- to at least be above the diagonal line from $(0,0)$ to $(1,1)$ and monotonic, i.e., never decreasing~\citep{scores1}. An ROC curve that is a straight line from $(0,0)$ to $(1,1)$ denotes a scoring function that is completely independent of $Y$, i.e., a trivial scoring function. 

\begin{figure}[t]
\centering
%\begin{center}
\includegraphics[width=0.85\columnwidth]{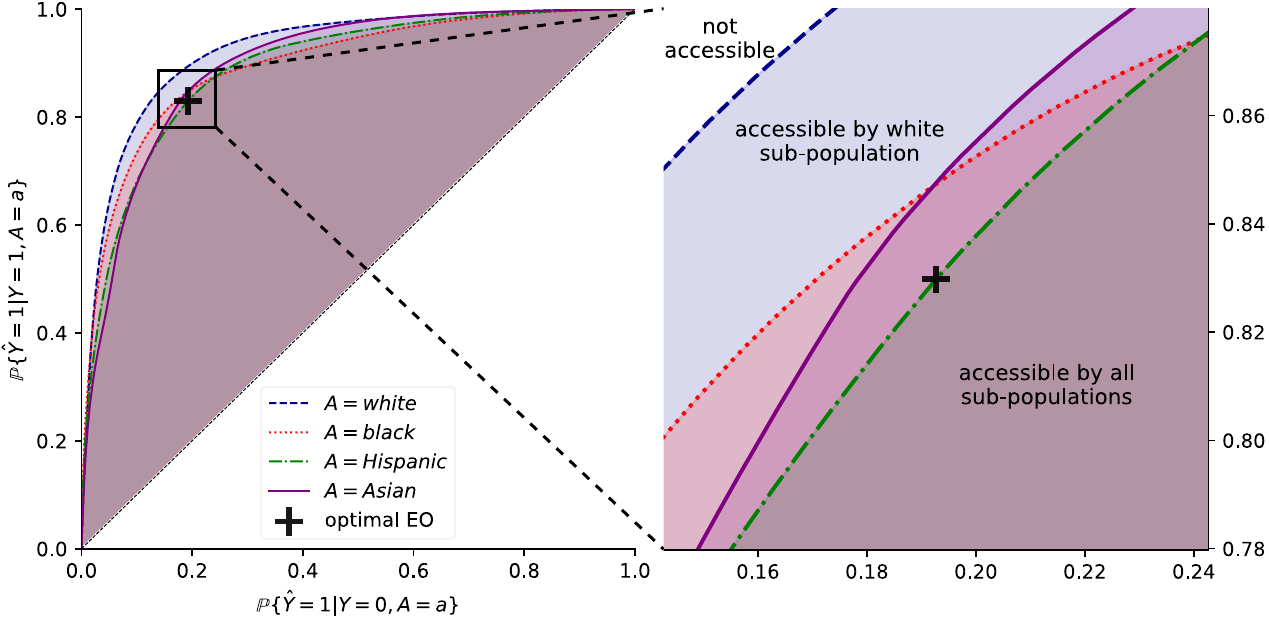}
\caption{ROC curves for the CreditRisk data set. The solution space for each (protected) group is given by all the points on their respective ROC curve when a single threshold is used. If we rely on multiple thresholds and randomisation, however, we expand the solution space to all the points on and below an ROC curve -- represented for each group as the coloured area. A fair solution, according to equalised odds, is any set of thresholds and probabilities such that each group achieves equal true and false positives.}
\label{ROC}
%\end{center}
\end{figure}

Optimising a scoring function $g$ with respect to complex definitions of fairness (such as equalised odds given by Definition~\ref{def::EO}) for multiple protected groups is more challenging than optimising $g$ for less strict fairness notions (e.g., demographic parity) due to the need to satisfy multiple constraints simultaneously. With post-processing, we assume that $g$ is \emph{fixed} and \emph{inaccessible}, i.e., a black box. We cannot therefore know or alter how scores are calculated from the input space, nor do we have access to the input space. This may be due to trade secrets or privacy concerns~\citep{risks9070132}, and applies to credit scoring~\citep{prop} among other domains. To achieve the desired notion(s) of fairness we therefore need to build an unbiased decision function $f_h$ upon $g$ by finding optimal thresholds for $h$ using \emph{only} the joint distributions of $R$, $A$ and $Y$.

When cardinality of the protected attribute $A$ is 2, optimal equalised odds can be achieved by fixing a single threshold at any point where the ROC curves are equal. If there are multiple points where the curves meet, the optimal solution (lowest false positive and highest true positive rates) is the intersection closest to $(0,1)$, i.e., the perfect model. Figure~\ref{ROC} shows the ROC curves stratified by a protected attribute $A$ (\emph{race}) for a loan repayment prediction based on credit scores $R$ from the CreditRisk data set.

The challenge arises when  the ROC curves do not touch or if $\vert A \vert > 2$. If the curves do not touch in $(0,1)\times(0,1)$, we can only satisfy equalised odds with a single threshold at the trivial points $(0,0)$ or $(1,1)$, i.e., assign the same outcome to all the scores. For $\vert A \vert > 2$ -- e.g., where $A=A_1\times A_2\times\cdots\times A_n$ may be a Cartesian product of $n$ protected characteristics -- it is highly unlikely for all the ROC curves to intersect at the same point (except for the trivial points). When using a single threshold, each group can only access false and true positive values that are \textbf{on their respective ROC curve} (shown in Figure~\ref{ROC} as the coloured curved lines). Using multiple thresholds and randomisation, however, allows each group to access all the points that are \textbf{below their respective ROC curve} and above the trivial scoring function (shown in Figure~\ref{ROC} as the coloured regions). The optimal point for equalised odds therefore becomes the point under all ROC curves that is closest to $(0,1)$.

%\subsection{Fixed Randomisation}
%\label{problems}
\citet{NIPS2016_9d268236} achieve equalised odds by setting group-specific thresholds $t_{y,a}$ -- where $t_{0,a}\leq t_{1,a}$, so $y\in\{0,1\}$ -- that are applied to the scoring function $g$. If a score falls between the thresholds designated for the protected group $a$, it is assigned a class at random with a probability given by the parameter $p_a\in\mathcal{I}$. Since thresholds are group-specific, we define a threshold-based classification function $h_a:\mathcal{R}\mapsto \mathcal{Y}$, where the probability of $h_a(r)=1$ is given by
\begin{equation}
\label{eq::h}
\zeta_a(r)=p_a\mathbbm{1}_{t_{0,a}}(r) + (1-p_a)\mathbbm{1}_{t_{1,a}}(r)\text{~,}
\end{equation}
for each protected sub-population $a$. In other words, $h_a(r)\sim B\big(1, \zeta_a(r)\big)$. We therefore define the final decision function as $f_{h_a} = h_a \circ g$, and Equation~\ref{eq::h} gives us
\begin{equation*}
    \mathbb{P}\{f_{h_a}(\mathbf{x})=1\vert A=a, X=\mathbf{x}\}= 
    \begin{cases}
        0 \quad &\text{if} \;\; g(\mathbf{x}) < t_{0,a} \\
        p_a \quad &\text{if} \;\; g(\mathbf{x}) \in [t_{0,a}, t_{1,a}) \\
        1 \quad &\text{if} \;\; g(\mathbf{x}) \geq t_{1,a}
    \end{cases}
        \text{~.}
\end{equation*}
We call this \textit{fixed randomisation}, as $r\in[t_{0,a}, t_{1,a})$ yields probability $p_a$ of $\widehat{Y}=1$. Setting $p_a = 0$, $p_a = 1$ or $t_{0,a} = t_{1,a}$ is synonymous to using a single threshold. A visual example of fixed randomisation is provided in Figure~\ref{multi-threshold}. 

Fixed randomisation is an effective approach to build a classifier $f_{h_a}$ based on a scoring function $g$ that satisfies group fairness such as equalised odds.
This strategy, however, exhibits a number of undesired properties;
most notably:
\begin{enumerate}[label=(\roman*)]
  \item\label{list:issues:1} it does not follow the general behaviour expected of a scoring function since all users who are subject to randomisation receive the same classification odds, no matter their score, but users whose scores are similar and near the thresholds are treated differently (ergo the example given in the introduction and shown in Figure~\ref{multi-threshold});
  \item\label{list:issues:2} even if $g$ is individually fair with well-defined $L_\mathcal{X}$, the discontinuities introduced by $\zeta_a$ at $t_{y,a}$ prevent $f_{h_a}$ from complying with individual fairness; and
  \item\label{list:issues:3} if $p_a\neq p_{a^\prime}$ then users from group $a$ cannot access the random classification odds offered to group $a^\prime$ and vice versa.
\end{enumerate}

Section~\ref{Introduction} has thoroughly demonstrated the adverse consequences of point~\ref{list:issues:1}. While users are made to believe that a higher score is better, e.g., their credit rating, fixed randomisation only exhibits this behaviour at the thresholds. Refer back to Figure~\ref{multi-threshold}, which shows that despite there being clear evidence of \emph{white} users with a credit score of $50$ being more likely to repay their loan than \emph{white} candidates whose credit score is $25$, both are equally likely (but not guaranteed) to receive a loan.

\begin{definition}[Classification Odds Distance]\label{def:prob_distance}
Given a decision function $h_a:\mathcal{R}\mapsto\mathcal{Y}$ such that $h_a(r)\sim B\big(1,\zeta_a(r)\big)$, we define the corresponding distance metric $d_\mathcal{Y}:\mathcal{I}\times\mathcal{I}\mapsto\mathcal{I}$ such that
\begin{equation*}
    d_\mathcal{Y}\big(h_a(r_1), h_a(r_2)\big) = \; \rvert \zeta_a(r_1) - \zeta_a(r_2)\lvert \qquad \forall r_1,r_2\in\mathcal{R}
        \text{~.}
\end{equation*}
Using Equation~\ref{eq::h}, the distance is the difference in odds of positive classification between two scores.
\end{definition}

Point~\ref{list:issues:2} concerns the classification behaviour around the thresholds $t_{y,a}$ and fixed randomisation parameter $p_a$, which create discontinuities in odds for the final decision function $f_{h_a}$.
To demonstrate this we use Definition~\ref{def:prob_distance}, which specifies a distance metric on the classification odds. 
Lipschitz conditions scale across compositions~\citep{federer1996geometric}, such that
\begin{equation*}
    d_\mathcal{Y}\big(h_a(g(\mathbf{x}_1)), h_a(g(\mathbf{x}_2))\big) 
    \leq
    L_\mathcal{R} d_\mathcal{R}\big(g(\mathbf{x}_1), g(\mathbf{x}_2)\big) 
    \leq
    L_\mathcal{R} L_\mathcal{X} d_\mathcal{X}(\mathbf{x}_1, \mathbf{x}_2) \qquad \forall \mathbf{x}_1,\mathbf{x}_2\in\mathcal{X}
        \text{~.}
\end{equation*}
Issues arise around the thresholds. Take
\begin{equation*}
r_1 = \lim_{z\to t_{y,a}^+} z \quad \text{and} \quad r_2 = \lim_{z\to t_{y,a}^-} z~\text{,}
\end{equation*}
so two scores approach a threshold from different sides. In such a case, from Equation~\ref{eq::h}, we have that
\begin{equation*}
    d_\mathcal{Y}\big(h_a(r_1), h_a(r_2)\big) = p_a \quad \text{or} \quad d_\mathcal{Y}\big(h_a(r_1), h_a(r_2)\big) = (1-p_a)
        \text{~,}
\end{equation*}
and $d_\mathcal{R}\big(r_1, r_2\big) \to 0$. Therefore
\begin{equation}
\label{explosion}
    \frac{d_\mathcal{Y}\big(h_a(r_1), h_a(r_2)\big)}{d_\mathcal{R}\big(r_1, r_2\big)}\to\infty \quad \text{and } \quad \frac{d_\mathcal{Y}\big(h_a(r_1), h_a(r_2)\big)}{d_\mathcal{R}\big(r_1, r_2\big)} 
    \leq 
    L_\mathcal{R}
        \text{~,}
\end{equation}
and thus $L_\mathcal{R}$ must be very large. As $r_1$ approaches $t_{y,a}$ from one side and $r_2$ from the other, $h_a$ is clearly not locally Lipschitz continuous since $d_\mathcal{R}\to0$ but $d_\mathcal{Y}\to p_a$ or $(1-p_a)$, one of which is always above $0$. In theory, $g$ could be crafted such that it cannot map individuals to values around the thresholds, however this would introduce discontinuities to $g$ and thus invalidate the Lipschitz condition. In this scenario, assuming $g$ satisfies the individual fairness constraint defined in Equation~\ref{lip}, $f_{h_a}$ must ultimately violate such an individual fairness constraint at the thresholds when fixed randomisation is employed. Fixed randomisation can therefore be seen as a \emph{step function} -- see Figure~\ref{multi-threshold} -- which is not uniformly continuous on any interval that contains $t_{y,a}$~\citep{stepdiscon}. Small changes can occur for a variety of reasons, e.g., a lack of instrumentation precision~\cite{threshold_effect} or noise due to human error~\citep{rudin2019stop}, and thus we argue that small changes should never dramatically change an individual's odds.

\begin{definition}[Equalised Individual Odds]\label{individualodds}
Given a probabilistic classifier $f_a:\mathcal{X}_a\mapsto \mathcal{Y}$, where $\mathcal{X}_a \subseteq \mathcal{X}\vert A=a$, defined by the probability curve $\zeta_a:\mathcal{R}\mapsto \mathcal{I}_a\subseteq\mathcal{I}$ such that $h_a(r)\sim B(1, \zeta_a(r))$ and $f_a = h_a\circ g \circ \cdots$, $f_a$ satisfies individual odds iff
\begin{equation*}
\exists 
r^\prime\in\mathcal{R}
\;\; \text{s.t.} \;\;
    \zeta_a(r)=\zeta_{a^\prime}(r^\prime) \quad \forall r\in\mathcal{R}
    \;\;
    \forall a,a^\prime\in A
        \;\;
        a\neq a^\prime
    \text{.}
\end{equation*}
Therefore, all sub-populations in $A$ must be capable of attaining classification odds available to all the other groups.
\end{definition}

Point~\ref{list:issues:3} highlights an interesting behaviour that gives rise to a novel, relatively weak, notion of fairness, which we call \emph{individual odds} -- see Definition~\ref{individualodds}.
To satisfy this fairness criterion
$\zeta_a$ does not necessarily need to be continuous but every point that it can reach must also be available to $\zeta_{a^\prime}$, so effectively we require $\mathcal{I}_a\equiv\mathcal{I}_{a^\prime}$.
Violating this constraint implies that there exists a subset of users from the $A=a$ sub-population that can never be treated the same as a portion of individuals from the $A=a^\prime$ group and vice versa.
Whenever $p_a\neq p_{a^\prime}$, the individual odds criterion is clearly not satisfied for fixed randomisation.
This definition of fairness bridges the, thus far somewhat separate, concepts of individual and group fairness as it considers the treatment of individual users in view of their assignment to distinct sub-populations determined by the protected attribute $A$.

\section{Constructing Curves for Preferential Randomisation}

Under these conditions, assuring group and individual fairness is equivalent to searching for solutions that are \emph{continuous} and \emph{smooth}, with a well-defined limit on $L_\mathcal{R}L_\mathcal{X}$, which also satisfy Definition~\ref{def::EO}. We therefore must find a combination of the group thresholds ($t_{0,a}$ and $ t_{1,a}$) and a curve between them that satisfies individual as well as group fairness.

\subsection{Defining Solution Behaviour}
\label{behaviour}

There are potentially infinite curves that satisfy the aforementioned conditions. In order to decrease the size of the solution space, we can impose further restrictions on the expected behaviour of the solution and parameterisation thereof. Where $h_a$ follows \emph{fixed randomisation}, we define \textit{preferential randomisation} as $z_a(r)\sim B\big(1, \phi_a(r)\big)$ to distinguish between the two; therefore, $f_{z_a} = z_a \circ g$ and
\begin{equation*}
    \mathbb{P}\{f_{z_a}(r)=1\vert A=a, R=r\}=\phi_a(r)~\text{.}
\end{equation*}
We expect preferential randomisation to behave as follows:
\begin{description}
    \item [Monotonicity]
    Larger values of $r$ should entail equal or higher chances of positive classification as argued by point~\ref{list:issues:1} in Section~\ref{postproc}, i.e.,
    \begin{equation*}
        \phi_a^\prime(r) \geq 0 \quad \forall r\in \mathcal{R}
        \text{~.}
    \end{equation*}
        \item [Continuity at boundaries]
    The solution should avoid sudden jumps in probability at the thresholds $t_{y,a}$ to satisfy point~\ref{list:issues:2}, i.e.,
    \begin{equation*}
        \phi_a(t_{y,a}) = y 
        \text{~.}
    \end{equation*}
    \item [Continuity for interval space]
    The curve that maps $\mathcal{R}$ to the classification probability must be well-defined at all points in $\mathcal{R}$ in compliance with point~\ref{list:issues:3}. If $\tilde{r}$ is any fixed point in $\mathcal{R}$,
    \begin{equation*}
        \lim_{r\to \tilde{r}^+}\phi_a(r) = \lim_{r\to \tilde{r}^-}\phi_a(r) \quad \forall \tilde{r} \in\mathcal{R}~\text{~,}
    \end{equation*}
where $r\to \tilde{r}^+$ is $r$ approaching $\tilde{r}$ from above and $r\to \tilde{r}^-$ is $r$ approaching $\tilde{r}$ from below.
\end{description}

Monotonicity between the thresholds guarantees that higher scores are treated better;
continuity within the interval ensures that the Lipschitz constant does not explode at the thresholds -- see Figure~\ref{postproc} for an example. Because no score can be outside of the $[\min(\mathcal{R}), \max(\mathcal{R})] = [\mathcal{R}_\alpha, \mathcal{R}_\omega]$ range, the output of $\phi_a$ does not need to span the entire probability range $[0,1]$ if the thresholds are fixed at the extremes, i.e., $t_{0, a} = \mathcal{R}_\alpha\implies\phi_a(t_{0,a}) \geq 0$ or $t_{1, a} = \mathcal{R}_\omega\implies\phi_a(t_{1,a}) \leq 1$. This is especially important when we can only access the final decisions $\mathcal{Y}$ as opposed to the scores $\mathcal{R}$, i.e., $g$ is a crisp classifier $g:\mathcal{X}\mapsto\mathcal{Y}$, in which case we require the ability to randomise the crisp predictions. With these constraints we can satisfy the requirements outlined in Section~\ref{postproc}.

\subsection{Viable Solutions from Linear Systems}
\label{systems}

Even with these constraints, the number of curves between each combination of thresholds that constitute viable solutions is still infinite. We therefore further constrict the solution space to piece-wise polynomials parameterised only by $t_{y,a}$ and $p_a$. We assume each solution takes the form
\begin{equation*}
    \psi_{a,y}(r)=v_y + b_yr + c_yr^2 +\cdots
    \text{~,}
\end{equation*}
and so for $\tau_a = t_{0,a} + (1-p_a)(t_{1,a} - t_{0,a})$,
%\begin{equation}
%\phi_a(r)=\mathbbm{1}_{[t_{0,a},\tau_a)}(r)\psi_{a, 0}(\frac{r-t_{0,a}}{t_{1,a}-t_{0,a}}) + \mathbbm{1}_{[\tau_a, t_{1,a})}(r)\psi_{a, 1}(\frac{r-t_{0,a}}{t_{1,a}-t_{0,a}}) + \mathbbm{1}_{[t_{1,a},\mathcal{R}_\omega]}(r)
%\end{equation}
\begin{equation}
\label{phicases}
    \phi_a(r) =
    \begin{cases}
        0 &\quad\text{if} \;\; r < t_{0,a} \\
        \psi_{a,0}(r) &\quad \text{if} \;\; r\in[t_{0,a}, \tau_a) \\
        \psi_{a,1}(r) &\quad \text{if} \;\; r\in [\tau_a, t_{1,a}) \\
        1 &\quad \text{if} \;\; r \geq t_{1,a}
    \end{cases}
        \text{~.}
\end{equation}
We choose this particular point of connection ($\tau_a$) because it ensures that all solutions (including $\zeta_a$) follow
\begin{equation*}
    \int_{t_{0,a}}^{t_{1,a}}\phi_a(r) dr = \int_{t_{0,a}}^{t_{1,a}} \zeta_a(r) dr \implies \int_\mathcal{R} \phi_a(r) dr = \int_\mathcal{R} \zeta_a(r) dr \text{~.}
\end{equation*}
This property guarantees that curves parameterised by the same thresholds and probabilities are comparable as they yield the same average probability between $t_{0,a}$ and $t_{1,a}$. The only difference between such solutions is their smoothness and continuity (see Appendix~\ref{avgprob_proof} for the proof). Finding families of closed-form solutions is achieved by using the continuity and monotonic constraints, with the addition of smoothness constraints as the order of the polynomial increases, and solving a full-rank linear system $M\mathbf{x}=\mathbf{b}$ (refer to Appendix~\ref{curves_proof} for details). 
Here, we consider four candidate curves:
\begin{description}
%\paragraph{Linear Form}
\item[linear form]
\begin{equation*}
    \begin{aligned}
        \psi_{a,0}(r) &= \frac{p_a(r - 1)}{1-p_a} \qquad\qquad \psi_{a,1}(r) &= \frac{(1-p_a)(r-1)}{p_a} + 1
    \end{aligned}
    %\label{eq::lin}
        \text{~,}
    \end{equation*}
    
%\paragraph{Quadratic Form}
\item[quadratic form]
    \begin{equation*}
    \begin{aligned}
        \psi_{a,0}(r) = \frac{p_a r^2}{(p_a-1)^2} \qquad\qquad \psi_{a,1}(r) = \frac{p_a^2+p_a-1}{p_a^2} - \frac{2(p_a-1)}{p_a^2}r + \frac{(p_a-1)}{p_a^2}r^2
        \end{aligned}
        %\label{eq::quad}
        \text{~,}
    \end{equation*}
    
%\paragraph{Cubic Form}
\item[cubic form]
    \begin{equation*}
    \begin{aligned}
        \psi_{a,0}(r) &= \frac{3p_ar^2}{p_a^2-2p_a+1} + \frac{2p_ar^3}{(p_a-1)^3} \\
        \psi_{a,1}(r) &= \frac{p_a^3+3p_a^2-3p_a+2}{p_a^3} +\frac{6(p_a^2-2p_a+1)}{p_a^3}r 
        + \frac{3(p_a^2-3p_a+2)}{p_a^3} r^2 + \frac{2(p_a-1)}{p_a^3} r^3
        \text{~, and}
        \end{aligned}
        %\label{eq::cub}
        \end{equation*}
        
%\paragraph{4\textsuperscript{th} Order Polynomial Form}
\item[4\textsuperscript{th} order polynomial form]
\begin{equation}
\begin{aligned}
    \psi_{a,y}(r) = (30p_a-12)r^2+(-60p_a+28)r^3+(30p_a-15)r^4
\end{aligned}
        \text{~.}
\label{eq::4}
\end{equation}
\end{description}
Their derivations are given in Appendix~\ref{curves_proof}. Note that the 4\textsuperscript{th} order polynomial (Equation~\ref{eq::4}) is not monotonic if $p_a\notin[\frac{2}{5},\frac{3}{5}]$.

\subsection{Validating Individual Fairness}
\label{sec::lip}

If $g$ is individually fair from the outset, validating that a given solution satisfies the individual fairness constraint is straight forward. From Definition~\ref{def:prob_distance} and Equation~\ref{explosion},
\begin{equation*}
     d_\mathcal{Y}\big(z_a(r_1), z_a(r_2)\big) 
    \leq 
    L_\mathcal{R} d_\mathcal{R}\big(r_1, r_2\big) \quad \implies \quad
    \frac{\lvert \phi_a(r_1) - \phi_a(r_2)\rvert}{\lvert r_1 - r_2 \rvert} \leq L_\mathcal{R}
    \text{~.}
\end{equation*}
Taking the limit $r_1\to r_2$, we get the definition of a derivative. Therefore, we can calculate $L_\mathcal{R}$ by considering the maximum derivative $L_\mathcal{R} = \max\big(\vert\phi^\prime_a(r)\rvert\big) \;\; \forall r\in[t_{0,a}, t_{1,a}]$ (proof in Appendix~\ref{lipcalc}). 
%\begin{equation*}
%    L_\mathcal{R} = \max\big(\vert\phi^\prime_a(r)\rvert\big) \quad \forall %r\in[t_{0,a}, t_{1,a}] 
%    \text{~.}
%\end{equation*}
Due to the definitions of each $\phi_a$, the maximum value of $\phi_a^\prime$ on $\mathcal{R}$ is always either at the thresholds or at the connection point, with the exception of the 4\textsuperscript{th} order polynomial for which $L_\mathcal{R}$ is where $\phi_a^{\prime\prime}(r)=0$ for $r\in (t_{0,a},t_{1,a})$, thus $L_\mathcal{R}$ is always known. Finding an optimal solution is therefore a case of identifying values of $t_{y,a}$ and $p_a$ for $\phi_a$ that satisfy Definition~\ref{def::EO} such that $L_\mathcal{R}L_\mathcal{X}$ is well-defined. While $L_\mathcal{R}$ is not guaranteed to be small, it is guaranteed to be finite. Taking the limit $p_a\to 1$ or $0$, $t_{0,a} \to t_{1,a}$, or $t_{1,a} \to t_{0,a}$, then $L_\mathcal{R}\to\infty$, which is synonymous with using a single threshold, hence invalidating equalised odds.
%\begin{equation*}
%    p_a\to 1 \; \textrm{or} \; 0 \textrm{,} \quad t_{0,a} \to t_{1,a}\text{,} \quad \textrm{or} \quad t_{1,a} \to t_{0,a} \quad \textrm{then} \quad L_\mathcal{R}\to\infty
%    \text{~,}
%\end{equation*}

\begin{figure}[t]
%\begin{center}
\centering
\includegraphics[width=0.85\columnwidth]{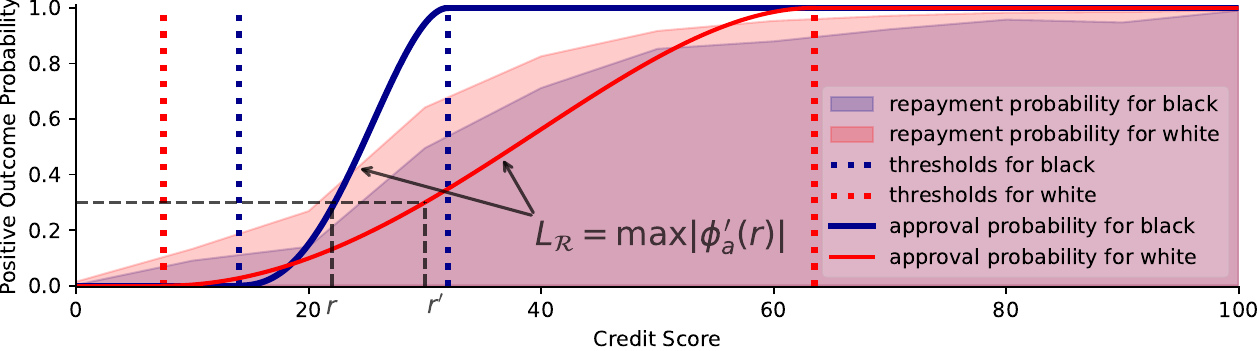}
\caption{Two-threshold preferential randomisation with smoothness constraints applied to probabilities (y-axis) output by a loan repayment classifier built upon credit scores (x-axis). It satisfies equalised odds for the protected attribute \textit{race} (black and white) by using it to assign approval probabilities.
This solution has \emph{no discontinuities} -- satisfying individual odds (Definition~\ref{individualodds}, see $r$ and $r^\prime$ for an example) and being $L_\mathcal{R}$ Lipschitz-continuous (Equation~\ref{lip}) -- and offers predictive performance marginally better than the fixed randomisation method shown in Figure~\ref{multi-threshold}.}
\label{multi-threshold_cont}
%\end{center}
\end{figure}

\section{Case Studies}
\label{casestudy}

Here we apply the method of preferential randomisation to two case studies: credit scoring for loan allocation (CreditRisk) and risk of recidivism (COMPAS). Source code for all the studies is available online\footnote{\url{https://github.com/teddyzander/McGIF}}.

\subsection{CreditRisk Case Study}
\label{sec:creditriskstudy}

To facilitate a direct comparison, we apply our method to the case study conducted by \citet{NIPS2016_9d268236}. Credit scores are often used to determine whether an individual should receive a loan or mortgage, to calculate interest rates and credit limits, and even to conduct background check on tenants~\citep{mortgage, HANSON201648}. The scoring function $g$ -- which calculates credit scores on input space $\mathcal{X}$ -- operates as a black box (see Section~\ref{postproc}), therefore we only observe the scores $R$ and cannot access $\mathcal{X}$ or $g$. 

The input space may contain attributes influenced by cultural background (i.e., related to race), possibly causing the joint distribution of $R$ and $Y$ to differ between sub-populations $A$. The CreditRisk data set captures the credit score's ability to predict defaulting on a loan (i.e., failing to repay it) for 90 days or more. The data show that as credit score increases, the likelihood of defaulting decreases (shown in Figure~\ref{credit_pmf} given in Appendix~\ref{ap:prob_func}) . The rate of these changes, however, is correlated with \emph{race}. Therefore, when a single threshold for each sub-population is optimised for maximum accuracy, the equalised odds (Definition~\ref{def::EO}) becomes $0.28$; we should strive for this fairness metric to be as close to $0$ as possible.

\begin{figure*}[b]
     \centering
     \begin{subfigure}[b]{0.28\textwidth}
         \centering
         \includegraphics[width=\textwidth]{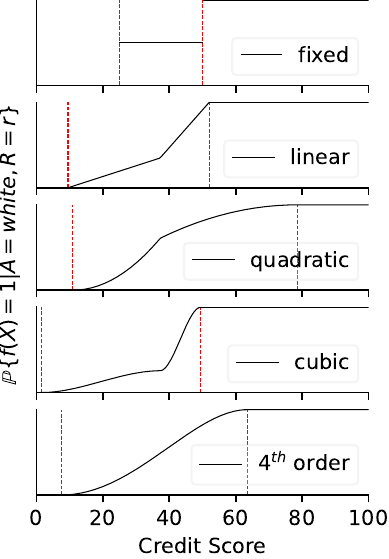}
     \end{subfigure}
     \hfill
     \begin{subfigure}[b]{0.28\textwidth}
         \centering
         \includegraphics[width=\textwidth]{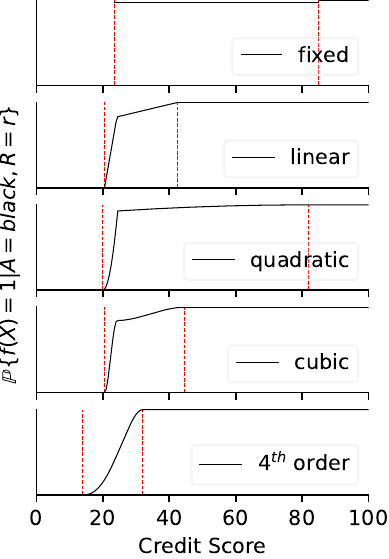}
     \end{subfigure}
     \hfill
     \begin{subfigure}[b]{0.28\textwidth}
         \centering
         \includegraphics[width=\textwidth]{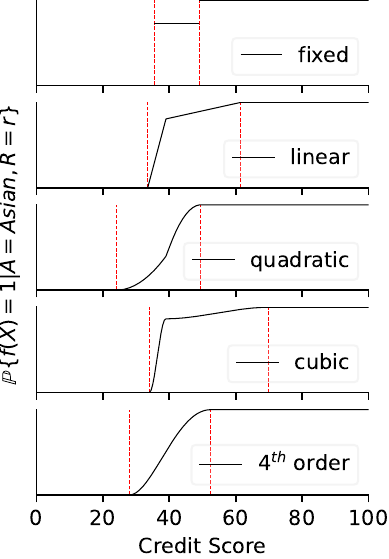}
     \end{subfigure}
     \caption{Probability curves corresponding to the results reported in Table~\ref{results}. All solutions have comparable accuracy and satisfy equalised odds but yield a different Lipschitz constant $L_\mathcal{R}$. The Hispanic group is omitted as it uses a single threshold $t_{y,a}=30$ (refer to Table~\ref{params} given in Appendix~\ref{ap:thresh}).}
     % for each value of the protected attribute \textit{race}
     \label{curves}
\end{figure*}

We overcome this by using different thresholds and probabilities (specified in Table~\ref{params} given in Appendix~\ref{ap:thresh}) achieved with a set of curves with differing smoothness constraints.
These curves honour the ``higher credit score leads to higher repayment probability'' dependency encoded in the underlying data. Referring back to the example introduced in Section~\ref{Introduction}, we can see from Figures~\ref{multi-threshold_cont} and~\ref{curves} that the \emph{white} user with a credit score of $49.5$ is now $2.6$--$5.25$ times more likely to receive a loan than the \emph{white} user with a credit score of $25$, depending on which continuous solution is chosen. The results -- reported in Table~\ref{results} -- show that the difference in accuracy and equalised odds between fixed randomisation and preferential randomisation is negligible (a change of $+0.016$ and $-0.000634$ respectively). The method additionally improves individual fairness by the Lipschitz constant on $\phi_a$ and through satisfying Definition~\ref{individualodds} (individual odds). Preferential randomisation can therefore be used to guarantee group and individual fairness through the notions of \emph{equalised odds} and \emph{individual odds}, and this encourages users to engage with the scoring model.

\begin{table*}[t]
\centering
%\small
\begin{tabular}{@{}r rrr c rrr c rrr @{}}
\toprule
       & \multicolumn{3}{c}{white} && \multicolumn{3}{c}{black} &&  \multicolumn{3}{c}{Asian} \\% Fairness
\cmidrule{2-4} \cmidrule{6-8} \cmidrule{10-12}
 \ & acc (\%) & EO & $L_\mathcal{R}$  &&  acc (\%) & EO & $L_\mathcal{R}$ &&  acc (\%) & EO & $L_\mathcal{R}$ \\% $G$
\midrule
fixed    & $82.424$ & $0.875$ & $\infty$ %~\citep{NIPS2016_9d268236}
&& $81.483$ & $0.382$ & $\infty$ 
&& $82.540$ & $\mathbf{0.175}$ & $\infty$ \\
linear      & $\mathbf{82.435}$ & $0.337$ & $0.044$
&& $81.480$ & $0.745$ & $0.222$
&& $82.544$ & $0.577$ & $0.148$ \\
quadratic    & $82.440$ & $1.336$ & $0.046$
&& $81.486$ & $0.558$ & $0.416$
&& $82.543$ & $0.347$ & $0.115$ \\
cubic    & $82.430$ & $\mathbf{0.241}$ & $0.091$
&& $\mathbf{81.487}$ & $\mathbf{0.103}$ & $0.338$
&& $82.542$ & $0.324$ & $0.265$ \\
4\textsuperscript{th} order & $82.432$ & $0.428$ & $\mathbf{0.027}$
&& $81.482$ & $0.569$ & $\mathbf{0.092}$
&& $\mathbf{82.545}$ & $0.554$ & $\mathbf{0.064}$ \\
\bottomrule
\end{tabular}
\caption{Accuracy (acc) as a percentage, equalised odds (EO) to the order of $\times 10^{-4}$, and Lipschitz constant ($L_\mathcal{R}$) per method for each value of the protected attribute \textit{race} in the CreditRisk loan repayment prediction task. The Hispanic group is not shown as it uses a single threshold ($t_{y,a}=30$) due to having the lowest ROC curve at the optimum, thus acting as the baseline for other \emph{races}.}
\label{results}
\end{table*}

\subsection{COMPAS Case Study\label{sec:compasstudy}}

\begin{figure}[t]
\centering
%\begin{center}
\includegraphics[width=0.85\columnwidth]{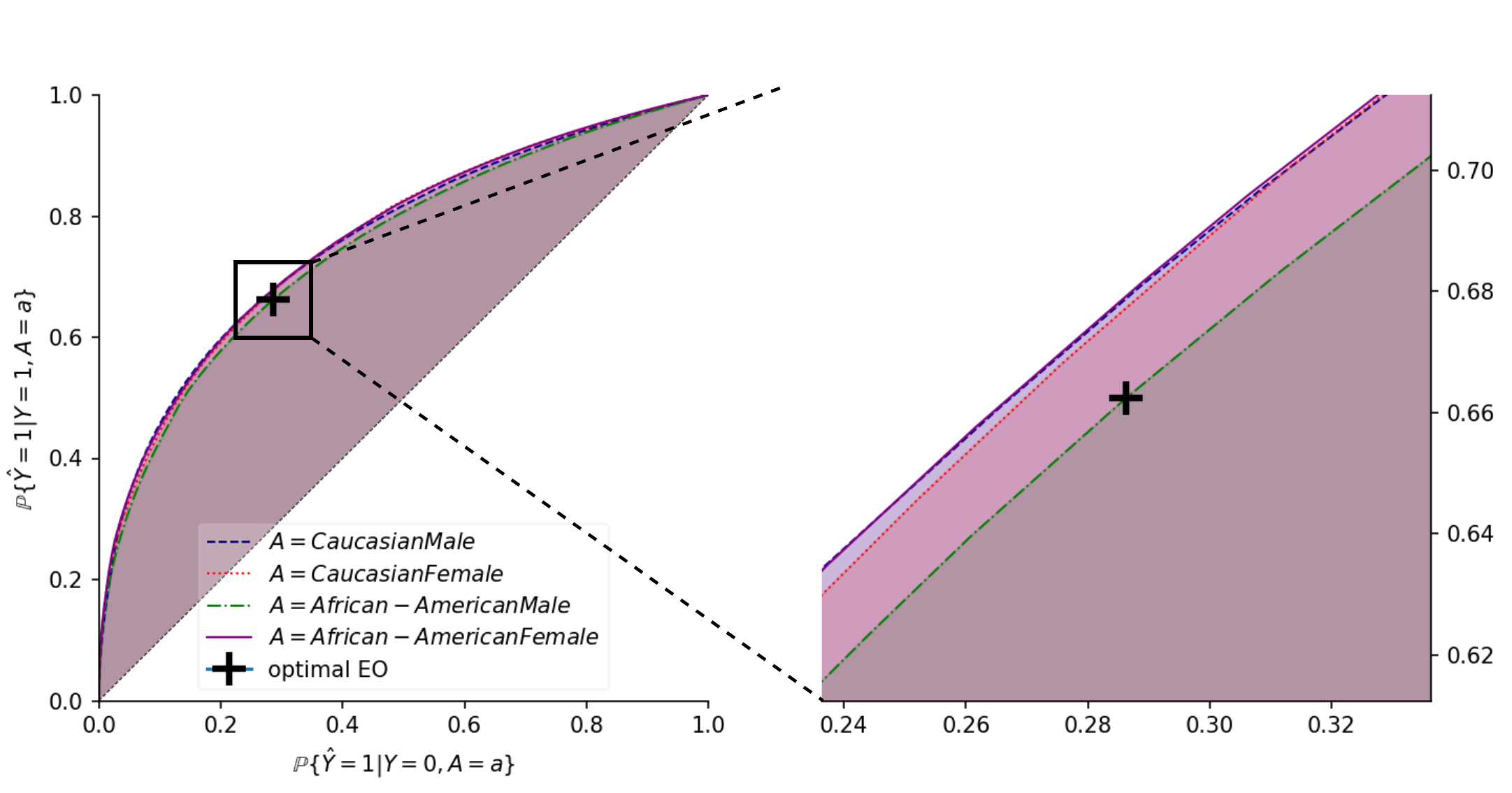}
\caption{ROC curves for the COMPAS data set. The coloured regions indicate areas accessible to each group. (Refer to Figure~\ref{ROC} for more details.)}
\label{ROC_compas}
%\end{center}
\end{figure}

The COMPAS (Correctional Offender Management Profiling for Alternative Sanctions) software%
\footnote{Refer to the COMPAS guide -- \url{https://www.equivant.com/practitioners-guide-to-compas-core} -- and documentation -- \url{https://doc.wi.gov/Pages/AboutDOC/COMPAS.aspx}.} %
is a commercial tool used across multiple U.S.\ states to analyse and predict a defendant's behaviour if released on bail. The software output can be considered by judges during sentencing, albeit such a practice must be disclosed. Specifically, COMPAS offers three insights: % though % declared % likelihood of re-offending
\begin{enumerate*}
    \item\label{compas:recidivism} likelihood of general recidivism (re-offending); % repeating an offence
    \item likelihood of violent recidivism (committing a violent crime); and
    \item likelihood of failing to appear in court (pretrial flight risk).
\end{enumerate*}
Here, we focus on the risk of re-offending using the raw COMPAS scores available in the ProPublica data set\footnote{\url{https://github.com/propublica/compas-analysis}}~\citep{angwin2016machine}. The COMPAS algorithm uses characteristics such as criminal history, known associates, drug involvement and indicators of juvenile delinquency in order to calculate a score $r$, where a higher score corresponds to a higher likelihood of recidivism. % the more likely an individual is to become a recidivist. % (\ref{compas:recidivism}), the general recidivism risk, % an individual's
As is the case with CreditRisk (Section~\ref{sec:creditriskstudy}), the scoring algorithm used by the COMPAS software is proprietary. Given its high stakes nature, it is important to understand the predictive behaviour of this tool since its social situatedness captured by the (protected) data features -- which are translated into the score $r$ -- may yield biased results~\citep{rudin2020age} as shown in Figure~\ref{ROC_compas}.

To this end, we define $A$ as the Cartesian product of two sensitive attributes -- \emph{sex} $A_1=\{\text{male}, \text{female}\}$ and \emph{race} $A_2=\{\text{Caucasian}, \text{African-American}\}$ -- found in the COMPAS data set, such that $A=A_1 \times A_2$ and so $\lvert A \rvert = 4$. Additionally, normalised COMPAS scores for a population of interest are denoted with $R$; the ground truth label for each score in $R$ is given by $Y$, where $1$ corresponds to individuals who committed an offence in a two-year time window; and $\widehat{Y}$ captures crisp predictions, with $1$ indicating high risk (of recidivism).
Studying the link between the scores and labels provided by the COMPAS data set -- refer to Figure~\ref{compas_pmf} given in Appendix~\ref{ap:prob_func} -- indicates that for most values of $r$ across all groups encoded by $A$ if $r_1 > r_2$, then $\mathbb{P}\{Y=1\vert R=r_1\} \geq \mathbb{P}\{Y=1\vert R=r_2\}.$ Therefore, we are in a good position to use the monotonic probability functions proposed in this paper to build the final classifier.
%\begin{equation}
%    \mathbb{P}\{Y=1\vert R=r_1\} \geq \mathbb{P}\{Y=1\vert R=r_2\} 
%    \text{~.}
%\end{equation}

\begin{table*}[t]
\centering
%\small
\begin{tabular}{@{}r rrr c rrr c rrr @{}}
\toprule
       & \multicolumn{3}{c}{Caucasian male} && \multicolumn{3}{c}{Caucasian female} &&  \multicolumn{3}{c}{African-American female} \\% Fairness
\cmidrule{2-4} \cmidrule{6-8} \cmidrule{10-12}
 \ & acc (\%) & EO & $L_\mathcal{R}$  &&  acc (\%) & EO & $L_\mathcal{R}$ &&  acc (\%) & EO & $L_\mathcal{R}$ \\% $G$
\midrule
fixed    & $\mathbf{67.400}$ & $2.412$ & $\infty$ %~\citep{NIPS2016_9d268236}
&& $\mathbf{67.595}$ & $1.413$ & $\infty$ 
&& $67.549$ & $1.625$ & $\infty$ \\
linear      & $67.384$ & $\mathbf{0.632}$ & $0.181$
&& $67.574$ & $1.190$ & $\mathbf{0.073}$
&& $67.555$ & $\mathbf{0.488}$ & $ 0.109$ \\
quadratic    & $67.386$ & $0.640$ & $0.254$
&& $67.594$ & $1.723$ & $0.129$
&& $67.548$ & $1.259$ & $0.214$ \\
cubic    & $67.395$ & $0.709$ & $0.254$
&& $67.589$ & $\mathbf{0.203}$ & $0.513$
&& $67.555$ & $2.349$ & $0.427$ \\
4\textsuperscript{th} order & $67.380$ & $1.145$ & $\mathbf{0.075}$
&& $67.576$ & $2.431$ & $0.080$
&& $\mathbf{67.574}$ & $1.863$ & $\mathbf{0.072}$ \\
\bottomrule
\end{tabular}
\caption{Accuracy (acc) as a percentage, equalised odds (EO) to the order of $\times 10^{-4}$, and Lipschitz constant ($L_\mathcal{R}$) per method for each value of the Cartesian product of the protected attributes \textit{race} and \textit{sex} in the COMPAS prediction task. The African-American male group is not shown as it uses a single threshold ($t_{y,a}=48$) due to having the lowest ROC curve at the optimum, thus acting as the baseline for other groups.}
\label{results_compas}
\end{table*}

\begin{figure*}[b]
     \centering
     \begin{subfigure}[b]{0.28\textwidth}
         \centering
         \includegraphics[width=\textwidth]{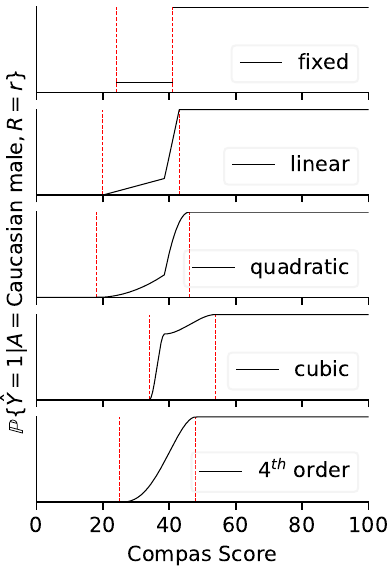}
     \end{subfigure}
     \hfill
     \begin{subfigure}[b]{0.28\textwidth}
         \centering
         \includegraphics[width=\textwidth]{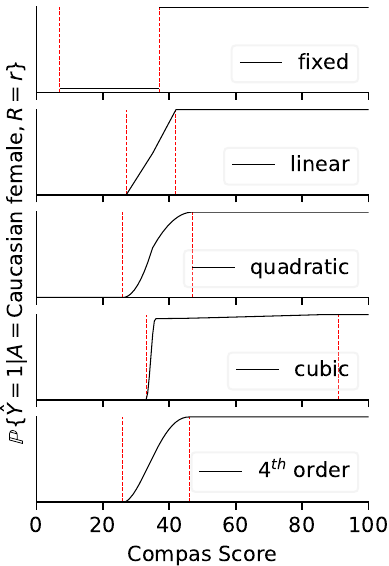}
     \end{subfigure}
     \hfill
     \begin{subfigure}[b]{0.28\textwidth}
         \centering
         \includegraphics[width=\textwidth]{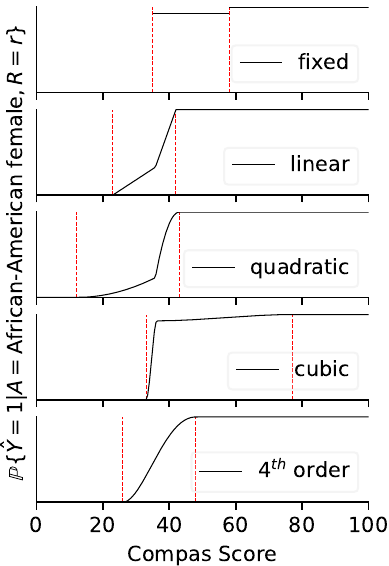}
     \end{subfigure}
     \caption{Probability curves corresponding to the results reported in Table~\ref{results_compas}. All solutions have comparable accuracy and satisfy equalised odds but yield a different Lipschitz constant $L_\mathcal{R}$. The African-American male group is omitted as it uses a single threshold $t_{y,a}=48$ (refer to Table~\ref{params_compas} given in Appendix~\ref{ap:thresh}).}
     % for each value of the protected attribute \textit{race}
     \label{curves_compas}
\end{figure*}

Given the aforementioned relationship, it is in the public's (and judicial system's) best interest to always increase the probability of classifying an individual as high-risk when the score $r$ increases. However, fixed randomisation does not allow for this. For example, under fixed randomisation a \emph{Caucasian male} with a COMPAS score in the $[24, 41)$ range has an 11.6\% chance of being classified as high-risk (see Table~\ref{params_compas} in Appendix~\ref{ap:thresh}); nonetheless, a \emph{Caucasian male} at the top of this score range is almost twice as likely to commit an offence as a \emph{Caucasian male} with a score at the low end of this range. Therefore, fixed randomisation is unfair on three fronts: %  since this more than likely corresponds to an increased chance of recidivism %  within the next two years than
\begin{enumerate*}
    \item \emph{Caucasian males} with scores in the lower range of $[24, 42)$ are treated the same as \emph{Caucasian males} with scores in the higher range of this interval;
    \item higher risk individuals are not labeled as such despite their scores indicating so; and % it is unfair to the general public to not label
    \item individuals whose outcome is randomised are never offered the same odds as members of other protected groups (in violation of Definition~\ref{individualodds}).
\end{enumerate*}
Notably, these arguments apply to all groups in the protected attribute $A$ and not only \emph{Caucasian males}. Small changes in score having a large impact on odds can have very real effects on individuals -- see the case of Mr.\ Rodriguez, whose analysis contained an error that caused his parole to be incorrectly denied~\citep{wexler2017computer} -- thus we argue that small changes should not dramatically change the chances of classification.

The between-group equalised odds measure when we maximise accuracy separately for each group is $0.148$. Mirroring Section~\ref{sec:creditriskstudy}, we apply our method to the COMPAS model in order to reduce the equalised odds disparity without breaking individual fairness. We therefore seek to calibrate the model with a combination of thresholds and probabilities that parameterise the continuous curves (defined in Section~\ref{systems}) using nothing but the joint distributions of $(R,A,Y)$. We then compare the continuous solutions to the step function solution (fixed randomisation) defined in Equation~\ref{eq::h}.
The results reported in Table~\ref{results_compas} and Figure~\ref{curves_compas} show that continuous curves can be used to simultaneously satisfy equalised odds, individual fairness and individual odds. Since low-scoring individuals are less likely to be classified as high-risk, defendants have an incentive to engage in behaviour that actively lowers their COMPAS score. Furthermore, public safety is prioritised more effectively since individuals with a measurably higher probability of recidivism are given higher odds of being classified as high-risk.

\section{Conclusion and Future Work}

In this work we demonstrated how using fixed randomisation to guarantee group fairness
may be detrimental to both the owners and users of a predictive model. Users with higher scores should be more likely to receive a better outcome -- a property that may be lost when enforcing group fairness. Ensuring this behaviour also allows the owners to preserve predictive performance and transparency of the automated decision-making process. By using the method proposed in this paper -- which relies on monotonic and continuous curves -- we can guarantee these properties. Our approach rewards building accurate scoring functions and adheres to the notion of individual fairness from the perspective of function composition. Importantly, the burden of accurate classification remains the sole responsibility of the model owner since our method forces all individuals to rely on the equalised odds measure of the worst-performing sub-population. This allocation of responsibility is desirable as owners can choose to invest in better predictors, data or scoring functions, whereas users in under-performing groups lack this agency.

Notably, our case study shows that there can exist multiple solutions that simultaneously satisfy equalised odds and individual fairness, which can be linked to \emph{model multiplicity}~\citep{sokol2022ethical}. When equalised odds, individual fairness and accuracy are comparable between groups, we can choose to discriminate the solutions based on other criteria. % against
Future work will explore this aspect of our curves;
specifically, we will consider:
\begin{enumerate}[label=(\arabic*)]
    \item the most robust curve for each group~\citep{ma2022on}; 
    \item curves such that $L_\mathcal{R}$ is closest between groups;
    \item the smoothest curves;
    \item curves that subject the fewest individuals to random outcomes, for example, $\min\lvert t_{1,a} - t_{0,a} \rvert \; \forall a\in A$ ; and
    \item curves that subject equal number of individuals to random outcomes between groups, e.g., $\min\sum_{\forall a\in A}\big(\lvert t_{1,a} - t_{0,a} \rvert - \lvert t_{1,a^\prime} - t_{0,a^\prime} \rvert\big)$ where $a\neq a^\prime$.
\end{enumerate}

\renewcommand{\acksname}{Acknowledgements}
\begin{acks}
    This research was conducted by the ARC Centre of Excellence for Automated Decision-Making and Society (project number CE200100005) funded by the Australian Government through the Australian Research Council.
\end{acks}

%%%%%%%%%%%%%%%%%%%%%%%%%%%%%%%%%%%%%%%%%%%%%%%%%%%%%%%%%%%%
\bibliographystyle{ACM-Reference-Format}
\bibliography{main}

\newpage
\appendix
\onecolumn

\section{Example Distance Functions}
\label{ap:dist}

If $\mathcal{M}$ is a metric space and $a,b,c\in\mathcal{M}$, then $d_\mathcal{M}:\mathcal{M}\times\mathcal{M}\mapsto \mathbb{R}^+$ and the following hold:
\begin{itemize}
    \item $d_\mathcal{M}(a,a) = 0$ -- the distance between a point and itself is $0$;
    \item if $a\neq b$, ~$d_\mathcal{M}(a,b) > 0$ -- the distance between two different points is strictly greater than $0$;
    \item $d_\mathcal{M}(a,b) = d_\mathcal{M}(b,a)$ -- the distance between two different points $a$ and $b$ is equal to the distance between $b$ and $a$; and
    \item $d_\mathcal{M}(a,c) \leq d_\mathcal{M}(a,b) + d_\mathcal{M}(b,c)$ -- the distance between any two points is equal to or less than the distance given by visiting another point on a journey between the original two points (triangle inequality).
\end{itemize}

\subsection{Euclidean Distance (Continuous Features)}
The $L^2$-norm is defined as
\begin{equation*}
    ||\mathbf{x}||_2 = \sqrt{\sum_{k=1}^N|x_k|^2}
    \text{~,}
    \label{l2}
\end{equation*}
and is the foundation of Euclidean distance $d_E:\mathcal{X}\times\mathcal{X}\mapsto\mathbb{R}$ defined as
\begin{equation*}
    d_E(\mathbf{x}_1, \mathbf{x}_2) = ||\mathbf{x}_1 - \mathbf{x}_2||_2
    \text{~,}
    \label{dl2}
\end{equation*}
and so
\begin{equation*}
    \begin{aligned}
        d_E(\mathbf{x}_1, \mathbf{x}_2) = \sqrt{\sum_{k=1}^N|x_{1,k} - x_{2,k}|^2}
        \text{~.}
    \end{aligned}
\end{equation*}

\subsection{Hamming Distance (Discrete Features)}
The $L^1$-norm is defined as
\begin{equation}
\label{l1norm}
    ||\mathbf{x}||_1 = \sum_{k=1}^N|x_k|
    \text{~,}
\end{equation}
and is the foundation of Hamming distance $d_H:\mathcal{X}\times\mathcal{X}\mapsto\{0,1,\ldots,N-1,N\}$, which counts the number of features that differ between two inputs $\mathbf{x}_1$ and $\mathbf{x}_2$, and is defined as
\begin{equation*}
    d_H(\mathbf{x}_1,\mathbf{x}_2) = ||\mathbf{x}_1 \oplus \mathbf{x}_2||_1
    \text{~,}
\end{equation*}
where $\oplus$ is the XOR operation. Therefore, $\mathbf{x}_1 \oplus \mathbf{x}_2$ is simply a vector of $0$'s and $1$'s such that
\begin{equation*}
%\label{xor}
    (\mathbf{x}_1 \oplus \mathbf{x}_2)_k = 
    \begin{cases}
        0 \quad &\text{if} \;\; x_{1,k} = x_{2,k} \\
        1 \quad &\text{if} \;\; x_{1,k} \neq x_{2,k}
    \end{cases}
        \text{~.}
\end{equation*}
For example,
\begin{equation*}
    \mathbf{x}_1 = \begin{bmatrix} 3\\1\\1\\1\\0\\1\\4\\2
    \end{bmatrix} \;\;
    \mathbf{x}_2 = \begin{bmatrix}
    3\\0\\2\\0\\0\\0\\4\\1
    \end{bmatrix}\implies
    \mathbf{x}_1 \oplus \mathbf{x}_2 = \begin{bmatrix}
    0\\1\\1\\1\\0\\1\\0\\1
    \end{bmatrix}
    \implies
    ||\mathbf{x}_1 \oplus \mathbf{x}_2||_1 = 5
    \text{~.}
\end{equation*}

\subsection{Gower's Distance (Mixed Continuous and Discrete Features)}

Take $\mathbf{x}_1, \mathbf{x}_2 \in \mathbb{R}^n$ that contain both continuous (numerical) variables and discrete (categorical) variables. We then consider each variable for $k=1,\ldots,n$. If the pair $x_{1,k}, x_{2,k}$ is continuous,
\begin{equation*}
    s_k = 1 - \frac{\lvert x_{1,k}-x_{2,k}\rvert}{V_n}
    \text{~,}
\end{equation*}
where $V_n$ is the range of the $k$\textsuperscript{th} feature. Fundamentally, the second term is the normalised $L^1$-norm (defined in Equation~\ref{l1norm}) on the differences between two vectors. However, if the pair $x_{1,k},x_{2,k}$ is discrete, we use the Iverson operation defined by
\begin{equation*}
    s_k = [x_{1,k} = x_{2,k}] =
    \begin{cases}
        0 \quad &\text{if} \;\; x_{1,k} \neq x_{2,k}\\
        1 \quad  &\text{if} \;\; x_{1,k} = x_{2,k}
    \end{cases}
        \text{~.}
\end{equation*}
As such, a value of $s_k=1$ for both continuous and discrete features implies that $x_{1,k}=x_{2,k}$, and $s_k=0$ implies that $x_{1,k}$ and $x_{2,k}$ are maximally different. We put this together to get Gower's Similarity Coefficient
\begin{equation}
\label{gowersim}
    S_G(\mathbf{x}_1,\mathbf{x}_2) = \frac{1}{n}\sum_{k=1}^n s_k
    \text{~,}
\end{equation}
which is bounded within $[0,1]$. However, this coefficient does not follow the axioms laid out at the beginning of this section as $S_G(a,a)=1$. Therefore, using Equation~\ref{gowersim} we define Gower's distance as
\begin{equation*}
    d_G(\mathbf{x}_1,\mathbf{x}_2) = \sqrt{1 - S_G(\mathbf{x}_1,\mathbf{x}_2)}
    \text{~,}
\end{equation*}
which offers the behaviour expected of a distance metric.

\section{Geometric Motivation for the Average Probability (Linear Case)}
\label{avgprob_proof}

We can parameterise each set of potential solutions for each value of $A$ (i.e., protected sub-population) by only three parameters -- $p_a$, $t_{0, a}$ and $t_{1, a}$ -- by constraining
the area under each curve as equal to $p_a(t_{1,a} - t_{0,a})$. This forces all potential solutions with the same set of parameters to have the same average probability between the thresholds.

\subsection{$\tau_a$ Proof (Point of Intersection)}
\label{tau_app}

\begin{figure}[t]
     \centering
     \centering
     \includegraphics[width=0.65\textwidth]{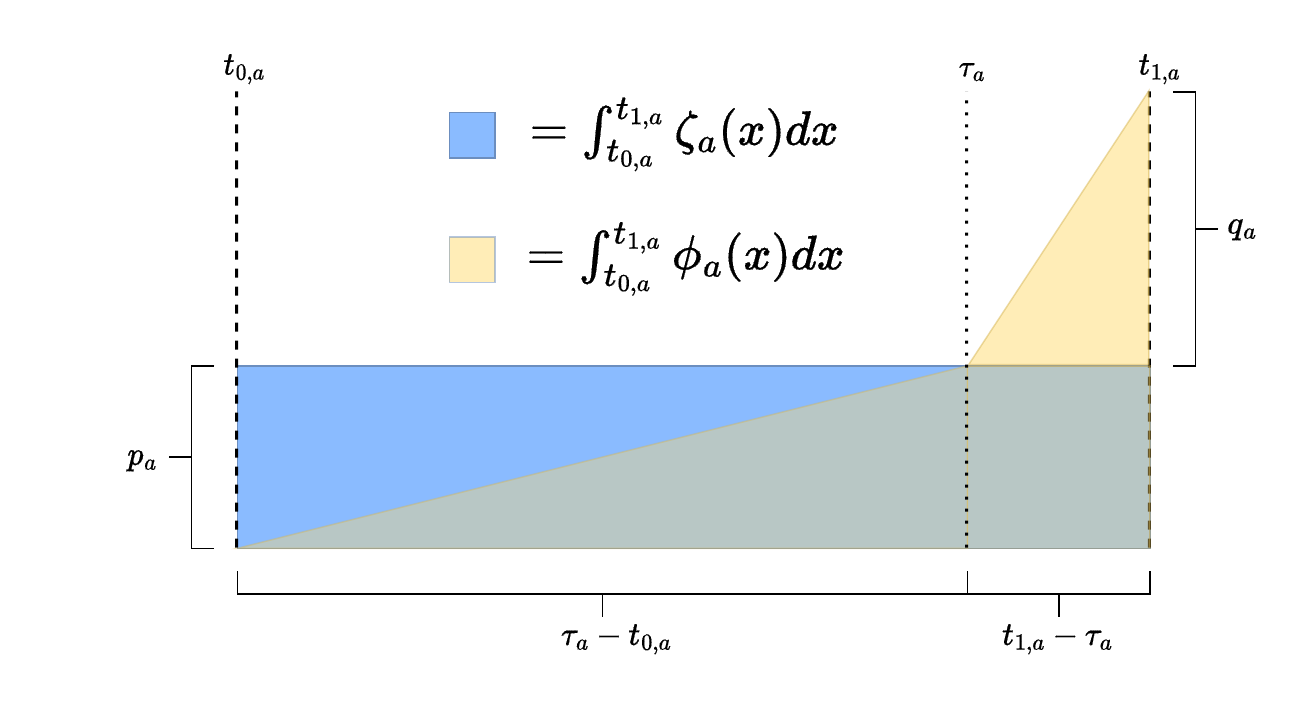}
     \caption{Geometric interpretation of a piece-wise linear solution using thresholds and probabilities. We can see that a step function yields a rectangular area -- blue space that denotes the average probability -- and is defined by the probability $p_a$ and the threshold interval size $t_{1,a}-t_{0,a}$. We expect this value to be equal to the area bounded by the piece-wise linear solution $\phi(x)$ and the x-axis (yellow space), which can be decomposed into simple geometric shapes and summed up.}
     \label{geom}
\end{figure}

Here, we discuss the details of bounding the piece-wise linear solution such that the two lines join at $\tau_a\in[t_{0,a},t_{1,a}]$. We present a proof that this value can be easily found, and is defined only through $q_a=1-p_a$, $t_{0,a}$ and $t_{1,a}$.

We begin by assuming that the solution is linear in nature and, as outlined in the paper (Section~\ref{behaviour}), it preserves the average probability of the step function within the interval $[t_{0,a},t_{1,a}]$. As such, we can generate a geometric interpretation of the solution as shown in Figure~\ref{geom}.
From here, we can see that finding $\tau_a$ is straight forward. By combining the area of a triangle equation, area of a rectangle equation, and forming an equality between equations, we get
\begin{equation*}
    p_a(t_{1,a}-t_{0,a})=\frac{p_a}{2}(\tau_a-t_{0,a}) + \frac{q_a}{2}(t_{1,a}-\tau_a) + (t_{1,a} - \tau_a)p_a
    \text{~,}
\end{equation*}
which can be interpreted as \emph{Rectangle = Triangle~1 + Triangle~2 + Small Rectangle}. We can then rearrange the notation as follows:
\begin{equation*}
    \begin{aligned}
        \frac{p_a}{2}(2T_{1,a} - 2T_{0,a} - \tau_a + t_{0,a} - 2T_{1,a} + 2\tau_a) &= \frac{q_a}{2}(t_{1,a} - \tau_a)
        \\
        \frac{p_a}{2}(\tau_a - t_{0,a}) &= \frac{q_a}{2}(t_{1,a} - \tau_a) \\
        p_a(\tau_a - t_{0,a}) &= q_a(t_{1,a} - \tau_a)
        \text{~.}
    \end{aligned}
\end{equation*}
Since we know that $p_a=1-q_a$,
\begin{equation*}
\begin{aligned}
    (1-q_a)(\tau_a - t_{0,a}) &= q_a(t_{1,a} - \tau_a) \\
    \tau_a - t_{0,a} - q_a (\tau_a - t_{0,a}) &= q_a(t_{1,a} - \tau_a) \\
    \tau_a - t_{0,a} &= q_a (\tau_a - \tau_a + t_{1,a} - t_{0,a}) \\
    \tau_a &= t_{0,a} + q_a (t_{1,a} - t_{0,a})
    \text{~.}
\end{aligned}
\end{equation*}
We therefore define $\Delta T_a = t_{1,a} - t_{0,a}$ to get the final result from Section~\ref{systems}:
\begin{equation}
    \tau_a = t_{0,a} + q_a \Delta T_a
    \text{~.}
    \label{tau}
\end{equation}
$\square$

\subsection{$\psi_1(x)$ and $\psi_0(x)$ Proof (Piece-wise Linear Solution)}
\label{linear}

We define the linear form of the interpolant as
\begin{equation}
    \phi_a(x) = 
    \begin{cases}
        0 \quad &x < t_{0,a} \\
        \psi_1(x) \quad & t_{0,a} \leq x < \tau_a \\ 
        \psi_0(x) \quad &\tau_a \leq x < t_{1,a}\\
        1 \quad & x \geq t_{1,a} 
        \label{interp}
    \end{cases}
        \text{~,}
\end{equation}
where each $\psi_n$ is linear, so
\begin{equation*}
    \psi_1(x) = vx+b \qquad \psi_0(x) = cx+d
    \text{~.}
\end{equation*}
In order to derive the final form, we must assume the following conditions (continuity):
\begin{enumerate}
\setlength\itemsep{0em}
    \item \label{ap:cond:1} $\psi_1(t_{0,a})=0$,
    \item \label{ap:cond:2} $\psi_0(t_{1,a})=1$, and
    \item \label{ap:cond:3} $\psi_1(\tau_a)=\psi_0(\tau_a)=p_a$.
\end{enumerate}
From conditions \ref{ap:cond:1} and \ref{ap:cond:3} we get
\begin{equation*}
    vT_{0,a} + b = 0 \qquad v\tau_a + b = p_a
    \text{~.}
\end{equation*}
From the difference of these equations we get
\begin{equation*}
    a(\tau_a-t_{0,a})=p_a
    \text{~.}
\end{equation*}
From the proof in Appendix~\ref{tau_app}, we know that $\tau_a = t_{0,a} + q_a \Delta T_a$ (Equation~\ref{tau}), giving
\begin{equation*}
    \begin{aligned}
    v(t_{0,a} + q_a \Delta T_a - t_{0,a}) &= p_a \\
    vq_a\Delta T_a &= p_a \\
    v &= \frac{p_a}{\Delta T_aq_a}
    \text{~.}
    \end{aligned}
\end{equation*}
Similarly, from conditions \ref{ap:cond:2} and \ref{ap:cond:3} we get
\begin{equation*}
    c t_{1,a} + d = 1 \qquad c \tau_a + d = p_a
    \text{~.}
\end{equation*}
The difference yields
\begin{equation*}
    c (t_{1,a} - \tau_a) = 1-p_a
    \text{~.}
\end{equation*}
From the definition of $\tau_a$ and $p_a$ we get
\begin{equation*}
    \begin{aligned}
        c(t_{1,a} - t_{0,a} - \Delta T_a q_a) &= q_a 
        \\
        c\Delta T_a (1 - q_a) &= q_a 
        \\
        c\Delta T_a p_a &= q_a \\
        c &= \frac{q_a}{p_a\Delta T_a}
    \text{~.}
    \end{aligned}
\end{equation*}
Substituting these values back into the equations yields the final parameters
\begin{equation*}
    \begin{aligned}
        -b &= \frac{p_a}{\Delta T_aq_a}t_{0,a} \qquad& -d &= \frac{q_a}{p_a\Delta T_a } t_{1,a} - 1 \\
        b &= - \frac{p_aT_{0,a}}{\Delta T_aq_a} \qquad&
        d &= 1 - \frac{q_aT_{1,a}}{p_a\Delta T_a}
    \text{~.}
    \end{aligned}
\end{equation*}
Putting it all back into the piece-wise equation and factorising then gives
\begin{equation}
    \phi_a(x) = 
    \begin{cases}
        0 \quad &x < t_{0,a} \\
        \frac{p_a}{\Delta T_aq_a}(x - t_{0,a}) \quad & t_{0,a} \leq x < \tau_a \\ 
        \frac{q_a}{\Delta T_ap_a}(x - t_{1,a}) + 1 \quad &\tau_a \leq x < t_{1,a} \\
        1 \quad & x \geq t_{1,a} 
        \label{finalinterp}
    \end{cases}
    \text{~.}
\end{equation}
$\square$

\subsection{Preservation of Average Probability Proof}

We previously stated that solutions with the same set of parameters should always have the same average probability. For example, the linear solution $\phi_a$ preserves group fairness introduced by the step function by maintaining the same predictive behaviour (on average) in the $[t_{0,a}, t_{1,a}]$ interval. This follows directly from how we defined
\begin{enumerate*}[label=(\arabic*)]
\setlength\itemsep{0em}
    \item the linear solution and
    \item the point of intersection.
\end{enumerate*}
Nonetheless, we can also prove this property directly. If $\zeta_a$ is the step function for $A=a$, we follow by stating that we require
\begin{equation*}
    \int_{-\infty}^{\infty} \zeta_a(x) dx = \int_{-\infty}^{\infty} \phi_a(x) dx
    \text{~.}
\end{equation*}
The first thing to observe is that
\begin{equation*}
    \begin{aligned}
        \zeta_a(x) &= \phi_a(x) \qquad &\textrm{if} \;\; x &< t_{0,a} \\
        \zeta_a(x) &= \phi_a(x)
        \qquad &\textrm{if} \;\; x &\geq t_{1,a}
    %\text{~,}
    \end{aligned}
\end{equation*}
from Equations~\ref{eq::h} and~\ref{phicases}, and so
\begin{equation*}
    \begin{aligned}
        \int_{-\infty}^{t_{0,a}} \zeta_a(x) dx &= \int_{-\infty}^{t_{0,a}} \phi_a(x) dx \\
        \int_{t_{1,a}}^{\infty} \zeta_a(x) dx &= \int_{t_{1,a}}^{\infty} \phi_a(x) dx
    \text{~.}
    \end{aligned}
\end{equation*}
We know that the integral in the interval for the step function is
\begin{equation*}
\begin{aligned}
    \int_{t_{0,a}}^{t_{1,a}} \zeta_a(x) dx &= \int_{t_{0,a}}^{t_{1,a}}p_a dx \\
    &= p_a(t_{1,a} - t_{0,a}) \\
    &= p_a\Delta T_a
    \text{~.}
    \end{aligned}
    %\label{avgprob}
\end{equation*}
From Equation~\ref{interp} in Appendix~\ref{linear} we know that
\begin{equation}
    \psi_1(x) = \frac{p_a}{\Delta T_aq_a}(x - t_{0,a}) \qquad
    \psi_0(x) = \frac{q_a}{\Delta T_ap_a}(x - t_{1,a}) + 1
    \text{~.}
    \label{psi}
\end{equation}
We can decompose the integral of the piece-wise linear solution into two integrals over the interval, so using Equation~\ref{psi} we get
\begin{equation}
    \int_{t_{0,a}}^{t_{1,a}} \phi_a(x)dx = \int_{t_{0,a}}^{\tau_a}\psi_1(x)dx + \int_{\tau_a}^{t_{0,a}}\psi_0(x)dx
    \text{~.}
    \label{psiint}
\end{equation}
Therefore, from Equation~\ref{finalinterp} we get
\begin{equation*}
    \begin{aligned}
        \int_{t_{0,a}}^{\tau_a}\psi_1(x)dx &= \int_{t_{0,a}}^{\tau_a} \frac{p_a}{\Delta T_aq_a}(x - t_{0,a}) dx \\
        & = \frac{p_a(\tau_a-t_{0a})^2}{2\Delta T_a q_a}
    \text{~.}
    \end{aligned}
\end{equation*}
From definition of $\Delta T_a$ and $\tau_a$ in Appendix~\ref{tau_app} we then get
\begin{equation}
    \frac{p_a(\tau_a-t_{0a})^2}{2\Delta T_a q_a} = \frac{(t_{1,a}-t_{0,a})p_aq_a}{2}
    \text{~.}
    \label{psi1}
\end{equation}
Also from Equation~\ref{finalinterp}, we have
\begin{equation*}
    \begin{aligned}
        \int_{\tau_a}^{t_{1,a}}\psi_0(x)dx &=
        \int_{\tau_a}^{t_{0,a}} \frac{q_a}{\Delta T_ap_a}(x - t_{1,a}) + 1 \\
        &= \frac{(t_{1,a}-\tau_a)(q_a\tau_a-t_{1,a}q_a+2\Delta T_a p_a)}{2\Delta T_a p_a}
    \text{~,}
    \end{aligned}
\end{equation*}
and again, from definition of $\Delta T_a$ and $\tau_a$ in Appendix~\ref{tau_app} we get
\begin{equation}
    \frac{(t_{1,a}-\tau_a)(q_a\tau_a-t_{1,a}q_a+2\Delta T_a p_a)}{2\Delta T_a p_a} = \frac{(t_{1,a} - t_{0,a})p_a(p_a+1)}{2}
    \text{~.}
    \label{psi2}
\end{equation}
Then, from Equations~\ref{psiint}, \ref{psi1} and \ref{psi2}, and recalling that $p_a+q_a=1$, we get
\begin{equation*}
    \begin{aligned}
        \int_{t_{0,a}}^{t_{1,a}} \phi_a(x)dx &= \frac{(t_{1,a}-t_{0,a})p_aq_a}{2} +\frac{(t_{1,a} - t_{0,a})p_a(p_a+1)}{2}\\
        &= \frac{(t_{1,a}-t_{0,a})}{2} (p_a(1-p_a)+p_a(1+p_a)) \\
        &= \frac{(t_{1,a}-t_{0,a})}{2}(p_a - p_a^2 + p_a + p_a^2)\\
        &=\frac{(t_{1,a}-t_{0,a})}{2}2p_a \\
        &= (t_{1,a}-t_{0,a}) p_a \\
        &= \Delta T_a p_a
    \text{~,}
    \end{aligned}
\end{equation*}
and therefore
\begin{equation*}
    \int_{t_{0,a}}^{t_{1,a}} \zeta_a(x)dx = \int_{t_{0,a}}^{t_{1,a}} \phi_a(x) dx
    \text{~,}
\end{equation*}
which means
\begin{equation*}
    \int_{-\infty}^{\infty} \zeta_a(x) dx = \int_{-\infty}^{\infty} \phi_a(x) dx
    \text{~.}
\end{equation*}
$\square$\\
Proofs for other curves follow the same logic.

\section{Obtaining Full-rank Linear Systems to Find Closed-form Piece-wise Solutions of Differing Smoothness}
\label{curves_proof}

\subsection{Linear System}

We search for a family of possible solutions for each group $\phi_a$, satisfying equalised odds (Definition~\ref{def::EO}), that adhere to the following constraints:
\begin{description}
    \item [continuity]
    \begin{equation*}
        \phi_a(t_{0,a})=0 \qquad \phi_a(t_{1,a})=1
        \text{~,}
    \end{equation*}
    \item [monotonicity]
    \begin{equation*}
        \phi_a^\prime(x)\geq0
        \text{~, and}
    \end{equation*}
    \item [preservation of probability]
    \begin{equation*}
        \int_{t_{0,a}}^{t_{1,a}}\phi_a(x) dx = \int_{t_{0,a}}^{t_{1,a}} p_a dx
        \text{~.}
    \end{equation*}
\end{description}
By assuming that each $\phi_{a,n}=a_n + b_nx^2 +\cdots$, we can use these constraints (as well as other, more strict constraints) to find solutions to this problem of varying smoothness by solving the linear system
\begin{equation*}
    A\mathbf{x}=b
    \text{~,}
\end{equation*}
where $\mathbf{x}=[a_0,b_0,\ldots, a_n,c_n, \ldots]^T$.
(Continuous, non-smooth solutions to this linear problem are given in Appendix~\ref{linear}.)

\subsection{Closed-form Smoothness for $p_a\in[\frac{2}{5},\frac{3}{5}]$}

Here, we search for a smooth closed-form solution to the above problem. For simplicity, we assume that $t_{0,a}=0$ and $t_{1,a}=1$, however the solution can be generalised to arbitrary thresholds by applying shift and stretch operations.

We have the following constraints:
\begin{enumerate}
\setlength\itemsep{0em}
    \item $\psi_a(0)=0$,
    \item $\psi_a(1)=1$,
    \item $\psi_a^\prime(0)=0$,
    \item $\psi_a^\prime(1)=0$, and
    \item $\int_{0}^{1}\psi_a(x) dx = \int_{0}^{1} p_a dx = p_a$.
\end{enumerate}
Having five constraints requires five coefficients, and so we assume that
\begin{equation*}
    \psi_a(x) = a_1 + b_1x + c_1x^2 +d_1x^3 +g_1x^4
    \text{~.}
\end{equation*}
We know that
\begin{equation*}
    \psi^\prime_a(x) = b_1 + 2c_1x + 3d_1x^2 + 4g_1x^3 \qquad \int_{0}^{1}\psi_a(x) dx = a_1 + \frac{1}{2}b_1 + \frac{1}{3}c_1 + \frac{1}{4}d_1 + \frac{1}{5}g_1
    \text{~,}
\end{equation*}
and so we have the following well-defined, full-rank linear system:
\begin{equation*}
    \begin{bmatrix}
    1 & 0 & 0 & 0 & 0 \\
    0 & 1 & 0 & 0 & 0 \\
    1 & 1 & 1 & 1 & 1 \\
    0 & 1 & 2 & 3 & 4 \\
    1 & \frac{1}{2} & \frac{1}{3} & \frac{1}{4} & \frac{1}{5} 
    \end{bmatrix}
    \begin{bmatrix}
        a_1 \\ b_1 \\ c_1 \\ d_1 \\ g_1
    \end{bmatrix}
    =
    \begin{bmatrix}
        0 \\ 1 \\ 0 \\ 0 \\ p_a
    \end{bmatrix}
    \text{~.}
\end{equation*}
After solving the system, we get
\begin{equation*}
    \psi_a(x) = (30p_a-12)x^2+(-60p_a+28)x^3+(30p_a-15)x^4
    \text{~.}
\end{equation*}
If the scoring function $f(\mathbf{x})=y$ and $y\in \mathbb{R}$, then this gives the final solution
\begin{equation*}
    \mathbb{P}\{\phi_a(y)=1\}=
    \begin{aligned}
    \begin{cases}
    0 \quad &\textrm{if} \;\; y < t_{0,a} \\
    \psi_a(\frac{y-t_{0,a}}{t_{1,a} - t_{0,a}}) &\textrm{if} \;\; t_{0,a} \leq y < t_{1,a} \\
    1 \quad &\textrm{if} \;\; y \geq t_{1,a}
    \end{cases}
    \text{~.}
    \end{aligned}
\end{equation*}
This function, however, is only monotonic for $p_a\in[\frac{2}{5},\frac{3}{5}]$.\\
$\square$

\subsection{Piece-wise Cubic Interpolant}

The closed-form, 4\textsuperscript{th} order solution satisfies the smoothness and boundary constraints, but violates the monotonic constraint for any probability outside of the $[\frac{2}{5},\frac{3}{5}]$ range. We can address this issue by constructing a piece-wise spline based on cubic polynomials given by
\begin{equation*}
    \psi_{a,n}(x) = a_n +b_n x + c_n x^2 + d_n x^3
    \text{~.}
\end{equation*}
However, we need to add two additional constraints to the optimisation problem:
\begin{enumerate*}[label=(\arabic*)]
\item an agreed meeting point and
\item an agreed derivative at the meeting point.
\end{enumerate*}
Again, assuming that $t_{0,a}=0$ and $t_{1,a}=1$ -- recall that we can shift and re-scale the solution later -- we get:
\begin{enumerate}
\setlength\itemsep{0em}
    \item $\psi_{a,0}(0)=0$,
    \item $\psi_{a,1}(1)=1$,
    \item $\psi_{a,0}^\prime(0)=0$,
    \item $\psi_{1,0}^\prime(1)=0$,
    \item $\psi_{a,0}(1-p_a)=p_a$,
    \item $\psi_{a,1}(1-p_a)=p_a$,
    \item $\psi_{a,0}^\prime(1-p_a) - \psi_{a,1}^\prime(1-p_a)=0$, and
    \item $\int_{0}^{1-p_a}\psi_{a,0}(x) dx +  \int_{1-p_a}^{1}\psi_{a,1}(x) dx = \int_{0}^{1} p_a dx = p_a$.
\end{enumerate}
Since
\begin{equation*}
    \psi_{a,n}^\prime(x) = b_n + 2c_n x + 3d_n x^2
    %\text{~,}
\end{equation*}
and
\begin{equation*}
\begin{aligned}
    \int_{0}^{1-p_a}\psi_{a,0}(x) dx =& a_0(1-p_a) + b_0\frac{(1-p_a)^2}{2} + c_0\frac{(1-p_a)^3}{3} + d_0\frac{(1-p_a)^4}{4} \\
    \int_{1-p_a}^{1}\psi_{a,1}(x) dx =& a_1(1-(1-p_a)) + b_0\frac{1-(1-p_a)^2}{2} + \\ &c_0\frac{1-(1-p_a)^3}{3} + d_0\frac{1-(1-p_a)^4}{4}
    \text{~,}
    \end{aligned}
\end{equation*}
we get the following linear system:
\begin{equation*}
    A = \begin{bmatrix}
        1 & 0 & 0 & 0 & 0 & 0 & 0 & 0 \\
        0 & 1 & 0 & 0 & 0 & 0 & 0 & 0 \\
        1 & 1-p_a & (1-p_a)^2 & (1-p_a)^3 & 0 & 0 & 0 & 0 \\
        0 & 0 & 0 & 0 & 1 & 1 & 1 & 1 \\
        0 & 0 & 0 & 0 & 0 & 1 & 2 & 3 \\
        0 & 0 & 0 & 0 & 1 & 1-p_a & (1-p_a)^2 & (1-p_a)^3 \\
        0 & 1 & 2(1-p_a) & 3(1-p_a)^2 & 0 & -1 & -2(1-p_a) & -3(1-p_a) \\
        1-p_a & \frac{(1-p_a)^2}{2} & \frac{(1-p_a)^3}{3} & \frac{(1-p_a)^4}{4} & 1-(1-p_a) & \frac{1-(1-p_a)^2}{2} & \frac{1-(1-p_a)^3}{3} & \frac{1-(1-p_a)^4}{4}
    \end{bmatrix}
\end{equation*}
\begin{equation*}
\mathbf{x} = 
\begin{bmatrix}
    a_0 \\ b_0 \\ c_0 \\ d_0 \\ a_1 \\ b_1 \\ c_1 \\ d_1
\end{bmatrix} \qquad\qquad
b=
    \begin{bmatrix}
        0 \\ 0 \\ p_a \\ 1 \\ 0 \\ p_a \\ 0 \\ p_a
    \end{bmatrix}
    \text{~.}
\end{equation*}
Solving $A\mathbf{x}=b$ yields
\begin{equation*}
\begin{aligned}
    &a_0 = 0 &b_0 = 0, \qquad \qquad \qquad \quad &c_0 = \frac{3p}{p_a^2-2p_a+p_a} &d_0 = \frac{2p_a}{(p_a-1)^3} \\
    &a_1 = \frac{p_a^3+3p^2-3p+2}{p_a^3} &b_1 = \frac{6(p_a^2-2p+1)}{p_a^3}\quad  &c_1 = \frac{3(p_a^2-3p+2)}{p_a^3} &d_1 = \frac{2(p_a-1)}{p_a^3}
    \text{~.}
\end{aligned}
\end{equation*}
We then take the solution to be
\begin{equation*}
    \mathbb{P}\{\phi_a(y)=1\}=
    \begin{aligned}
    \begin{cases}
    0 \quad &\textrm{if} \;\; y < t_{0,a} \\
    \psi_{a,0}(\frac{y-t_{0,a}}{t_{1,a} - t_{0,a}}) &\textrm{if} \;\; t_{0,a} \leq y < \tau_a \\
    \psi_{a,1}(\frac{y-t_{0,a}}{t_{1,a} - t_{0,a}}) &\textrm{if} \;\; \tau_a \leq y < t_{1,a}  \\
    1 \quad &\textrm{if} \;\; y \geq t_{1,a}
    \end{cases}
    \text{~.}
    \end{aligned}
\end{equation*}
$\square$

\section{Maximum $L_\mathcal{R}$ for Different Curves}
\label{lipcalc}

As discussed in Section~\ref{systems}, we know that
\begin{equation*}
    L_\mathcal{R}=\max\lvert \phi_a^\prime(r) \rvert
    \text{~.}
\end{equation*}
Since these curves are specified through closed-form solutions parameterised by $t_{a,y}$ and $p_a$ on a known interval $\mathcal{R}$, $L_\mathcal{R}$ can be found analytically for each curve. Here, we show the derivation procedure for the linear and 4\textsuperscript{th} order solutions. The other curves (cubic and quadratic) follow the same protocol.

The linear solution is defined in Equation~\ref{finalinterp}. As such, we know that
\begin{equation*}
    \phi_a^\prime(r) =
    \begin{cases}
        0 \quad &r \geq t_{1,a} \\
        \frac{p_a}{\Delta T_a (1-p_a)} \quad &\tau_a \leq r < t_{1,a} \\
        \frac{1-p_a}{\Delta T_a p_a} \quad &t_{0,a} \leq r < \tau_a \\
        0 \quad &r<t_{0,a}
    \end{cases}
        \text{~.}
\end{equation*}
Thus, the value of $L_\mathcal{R}$ is related to the value of $p_a$ and the distance between the thresholds with
\begin{equation*}
    \max\lvert \phi_a^\prime(r) \rvert = 
    \begin{cases}
    \frac{1}{\Delta T_a} \quad &\textrm{if } p_a=\frac{1}{2}\\
        \frac{p_a}{\Delta T_a (1-p_a)} \quad &\textrm{if } p_a > \frac{1}{2} \\
        \frac{1-p_a}{\Delta T_a p_a} \quad &\textrm{if } p_a < \frac{1}{2}
    \end{cases}
        \text{~.}
\end{equation*}
The 4\textsuperscript{th} order is define in Equation~\ref{eq::4}, and takes the form
\begin{equation*}
    \phi_a(r) = 
    \begin{cases}
        1 \quad &r \geq t_{1,a} \\
        \psi_{a}(\frac{r-t_{0,a}}{t_{1,a} - t_{0,a}})
         &t_{0,a} \leq r < t_{1,a} \\
        0 \quad &r < t_{0,a}
    \end{cases}
        \text{~,}
\end{equation*}
where
\begin{equation}
    \psi_{a}(x) = (30p_a-12)x^2+(-60p_a+28)x^3+(30p_a-15)x^4
    \text{~.}
    \label{smoothness}
\end{equation}
Since the definition of $\phi_a(r)$ is always constrained such that it is monotonic and $\phi_a(t_{y,a})=y$, the maximum derivative always occurs at the point of inflexion, or
\begin{equation*}
    \psi_{a}^{\prime\prime}(\frac{r-t_{0,a}}{t_{1,a} - t_{0,a}}) = 0
    \text{~.}
\end{equation*}
From Equation~\ref{smoothness}:
\begin{equation*}
    \psi_{a}^{\prime\prime}(x) = 12(30p_a-15)x^2+6(-60p_a+28)x+2(30p_a-12)
    \text{~,}
\end{equation*}
and so from the quadratic formula for $\psi_{a}^{\prime\prime}(x)=0$ we get
\begin{equation*}
\begin{aligned}
    x^\star &= \frac{-(6(-60p_a+28)) - \sqrt{(6(-60p_a+28))^2 - 4(12(30p_a-15))(2(30p_a-12))}}{2(12(30p_a-15))} \\
    & = -\frac{7-15p_a+\sqrt{75p_a^2-75p_a+19}}{30p_a-15}
    \text{~,}
    \end{aligned}
\end{equation*}
where we ignore the second root, as it takes $x^\star$ out of range. We can use this to find $\max\lvert \phi_a^\prime(x)\rvert = \phi_a^\prime(x^\star)$, which assumes $t_{0,a}$ and $t_{1,a}$ are fixed at $0$ and $1$ respectively, and apply a re-scaling such that
\begin{equation*}
    \max\lvert \phi^\prime_a(r) \rvert = \frac{\lvert \phi_a'(x^\star)\rvert}{t_{1,a} - t_{0,a}}
    \text{~.}
\end{equation*}
This method works for all valid values of $p_a$, except when $p_a=\frac{1}{2}$ where $\psi_a(x)$ is reduced to a cubic equation, and thus the point of inflexion is perfectly between the two thresholds:
\begin{equation*}
    \max\lvert \phi^\prime_a(r) \rvert = \phi_a^\prime \big(t_{0,a} + \frac{1}{2}(t_{1,a}-t_{0,a})\big)\text{~,}
\end{equation*}
%Then:
in which case
\begin{equation*}
    \mathcal{L}_\mathcal{R} = \max\lvert \phi^\prime_a(r) \rvert\text{~.}
\end{equation*}

\section{Probability Functions}
\label{ap:prob_func}

\begin{figure}[!h]
\centering
%\begin{center}
\includegraphics[width=0.75\columnwidth]{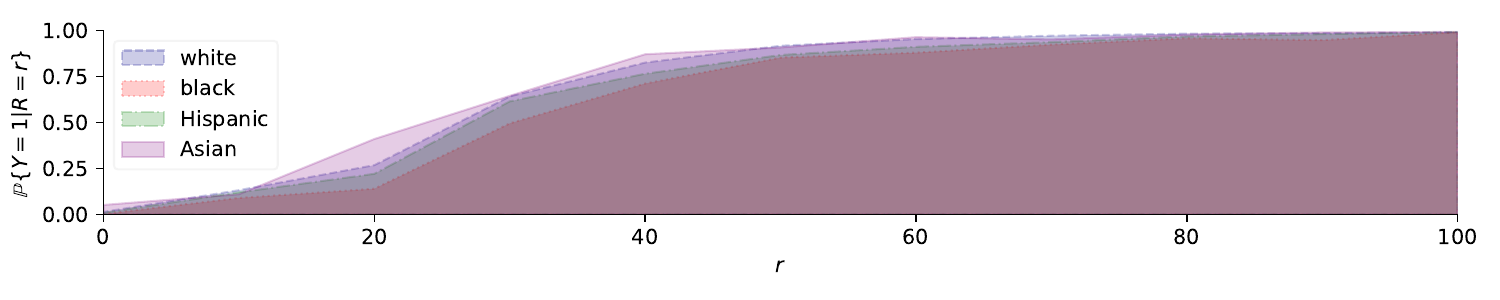}
\caption{Probability function for the CreditRisk data set. An increase in credit score $r$ (x-axis) generally leads to an increase in the probability of an individual not defaulting on a loan (y-axis) in the last 90 days ($Y=1$).}
\label{credit_pmf}
%\end{center}
\end{figure}

\begin{figure}[h!]
\centering
%\begin{center}
\includegraphics[width=0.75\columnwidth]{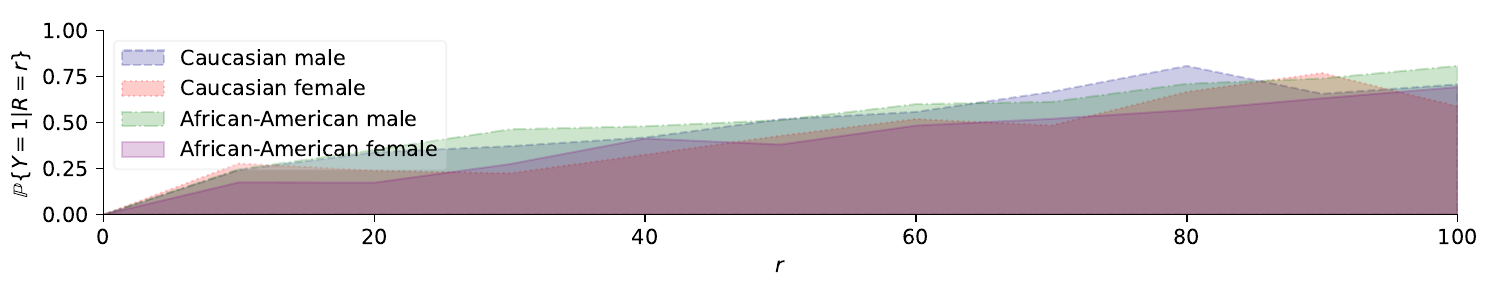}
\caption{Probability function for the COMPAS data set. An increase in score $r$ (x-axis) generally leads to an increase in the probability of committing an offence (y-axis) in a two-year time window ($Y=1$).}
\label{compas_pmf}
%\end{center}
\end{figure}

%\afterpage{\clearpage}
\FloatBarrier

\section{Optimal Thresholds for Fairness}
\label{ap:thresh}

\begin{table*}[h!]
\centering%\small%\scriptsize% \ra{1.3}
%\small
\begin{tabular}{@{}r rrr c rrr c rrr c rrr@{}}
\toprule
       & \multicolumn{3}{c}{white} && \multicolumn{3}{c}{black} && \multicolumn{3}{c}{Hispanic} && \multicolumn{3}{c}{Asian} \\% Fairness
\cmidrule{2-4} \cmidrule{6-8} \cmidrule{10-12} \cmidrule{14-16}
 & $t_{0,a}$\ & $t_{1,a}$ & $p_{a}$ && $t_{0,a}$\ & $t_{1,a}$ & $p_{a}$ && $t_{0,a}$\ & $t_{1,a}$ & $p_{a}$ && $t_{0,a}$\ & $t_{1,a}$ & $p_{a}$ \\% $G$
\midrule
fixed    & $25.0$    & $50.0$      &   $0.500$    && %~\citep{NIPS2016_9d268236}
$23.5$    & $85.0$      &   $0.972$    && 
$30.0$    & $30.0$      &   $0.000$    && 
$35.5$    & $49.0$      &   $0.728$    \\
linear    & $9.5$    & $52.0$      &   $0.348$  &&
$20.5$    & $42.5$      &   $0.830$    && 
$30.0$    & $30.0$      &   $0.000$    && 
$33.5$    & $61.5$      &   $0.806$    \\
quadratic    & $11.0$    & $78.5$      &   $0.610$    &&
$20.0$    & $82.0$      &   $0.928$    &&
$30.0$    & $30.0$      &   $0.000$    &&
$24.0$    & $49.5$      &   $0.406$    \\
cubic    & $1.5$    & $49.5$      &   $0.256$    &&
$20.5$    & $44.5$      &   $0.844$    &&
$30.0$    & $30.0$      &   $0.000$    &&
$34.0$    & $70.0$      &   $0.864$    \\
4\textsuperscript{th} order    & $7.5$    & $63.5$      &   $0.468$    && 
$14.0$    & $32.0$      &   $0.426$    && 
$30.0$    & $30.0$      &   $0.000$    &&
$28.0$    & $52.5$      &   $0.546$    \\
\bottomrule
\end{tabular}
\caption{Thresholds and probabilities for each curve across all classes of the CreditRisk data set. See Figure~\ref{curves} for visualisation.}
\label{params}
\end{table*}

\begin{table*}[!h]
\centering%\scriptsize% \ra{1.3}
%\small
\begin{tabular}{@{}r rrr c rrr c rrr c rrr@{}}
\toprule
       & \multicolumn{3}{c}{Caucasian male} && \multicolumn{3}{c}{Caucasian female} && \multicolumn{3}{c}{African-American male} && \multicolumn{3}{c}{\makesamewidth[c]{African-American male}{African-American female}} \\% Fairness
\cmidrule{2-4} \cmidrule{6-8} \cmidrule{10-12} \cmidrule{14-16}
 & $t_{0,a}$\ & $t_{1,a}$ & $p_{a}$ && $t_{0,a}$\ & $t_{1,a}$ & $p_{a}$ && $t_{0,a}$\ & $t_{1,a}$ & $p_{a}$ && $t_{0,a}$\ & $t_{1,a}$ & $p_{a}$ \\% $G$
\midrule
fixed    & $24.0$    & $41.0$      &   $0.116$    &&  %~\citep{NIPS2016_9d268236}
$7.0$    & $37.0$      &   $0.048$    && 
$48.0$    & $48.0$      &   $0.000$    && 
$35.0$    & $58.0$      &   $0.930$    \\
linear    & $20.0$    & $43.0$      &   $0.194$  &&
$27.0$    & $42.0$      &   $0.478$    && 
$48.0$    & $48.0$      &   $0.000$    && 
$23.0$    & $42.0$      &   $0.326$    \\
quadratic    & $18.0$    & $46.0$      &   $0.266$    &&
$26.0$    & $47.0$      &   $0.576$    &&
$48.0$    & $48.0$      &   $0.000$    &&
$12.0$    & $43.0$      &   $0.232$    \\
cubic    & $34.0$    & $54.0$      &   $0.772$    &&
$33.0$    & $91.0$      &   $0.952$    &&
$30.0$    & $30.0$      &   $0.000$    &&
$33.0$    & $77.0$      &   $0.926$    \\
4\textsuperscript{th} order    & $25.0$    & $48.0$      &   $0.412$    && 
$26.0$    & $46.0$      &   $0.557$    && 
$48.0$    & $48.0$      &   $0.000$    &&
$26.0$    & $48.0$      &   $0.550$    \\
\bottomrule
\end{tabular}
\caption{Thresholds and probabilities for each curve across all classes of the COMPAS data set. See Figure~\ref{curves_compas} for visualisation.}
\label{params_compas}
\end{table*}

\end{document}